\documentclass[11pt]{article}
\usepackage[preprint]{acl}
\usepackage{times}
\usepackage[T1]{fontenc}
\usepackage{microtype}
\usepackage{amsmath,amssymb}
\usepackage{booktabs}
\usepackage{enumitem}
\usepackage{graphicx}
\usepackage{xcolor}
\usepackage{tikz}
\usetikzlibrary{positioning,arrows.meta}
\definecolor{steel}{RGB}{0,114,178}
\definecolor{leaf}{RGB}{0,120,95}
\newcommand{\Binferred}{B_{\mathrm{inf}}}
\newcommand{\Bassumed}{B_{\mathrm{assume}}}
\newcommand{\Ifilt}{I_{\mathrm{filtered}}}

\title{Can LLMs Give Good Product Advice?\\
A Framework for Diagnosing Failures}
\author{
Assaf Hallak\textsuperscript{1}\quad
Yftah Ziser\textsuperscript{1,2}\quad
Gal Chechik\textsuperscript{1,3}\\
\textsuperscript{1}NVIDIA Research \\
\textsuperscript{2}University of Groningen\quad
\textsuperscript{3}Bar-Ilan University \\
\texttt{ahallak@nvidia.com}
}
\date{}
\hypersetup{pdfauthor={Assaf Hallak, Yftah Ziser, Gal Chechik}}

\begin{document}
\maketitle

\begin{abstract}
Evaluating product advice from a language model requires assessing
whether the recommendation meets the user's needs and identifying
where the model falls short when it does not.
However, matching a product mentioned in an online reply cannot
answer either question on its own: several products may be suitable,
and the recorded answer may itself be poor advice.
 When shown a community-listed alternative alongside
their own sampled car recommendation, six models prefer the alternative
in 1.9--6.5\% of eligible disagreements, consistently across both
source-blind presentation orders.
We propose a diagnostic framework that separates the user's stated
needs, the model's added assumptions, candidate generation, filtering,
and final choice, with 19 possible failure modes and the evidence
needed to assess them.
We apply this framework to product-advice data and to a
separate product-selection task with externally assigned relevance
labels. The experiments distinguish naming a community-listed product from
selecting it and show how answer extraction and candidate-list
construction affect the outcome. 
\end{abstract}

\section{Introduction}
\label{sec:intro}

Consider a parent asking a language model for a family-car
recommendation. The car must seat six people and cost at most
\$20,000; either an SUV or a minivan is acceptable, and lower running
costs are preferred. Recommending a five-seat car is a clear error;
choosing one suitable minivan over another may be a reasonable
trade-off. The task is to give advice that
fits the request, not necessarily to name one uniquely correct car.

Our goal is to answer two questions about product advice from a
large language model (LLM): \emph{can it give objectively good advice,
and where are its weaknesses when it falls short?} By good advice we
mean a recommendation that satisfies the user's firm requirements,
rests on accurate product facts, and makes a defensible trade-off
among the user's preferences. Seating capacity can be checked directly;
other judgments depend on priorities that may leave several acceptable choices.

\begin{figure*}[t]
\centering
\begin{tikzpicture}[
  font=\small,
  box/.style={draw=steel,fill=steel!5,rounded corners=2pt,
    align=left,text width=25mm,inner sep=4pt,minimum height=20mm},
  arrow/.style={-{Stealth[length=2mm]},draw=steel},
  node distance=4mm]
\node[box] (q) {\textbf{Request}\\Budget; use case;\\requirements;\\trade-offs};
\node[box,right=of q] (f) {\textbf{Evidence}\\$F$: stated facts\\$\Binferred$: inference\\$\Bassumed$: prior};
\node[box,right=of f] (c) {\textbf{Criteria}\\$C$: hard limits\\$P$: preferences\\and priorities};
\node[box,right=of c] (i) {\textbf{Options}\\$I$: candidates\\$\Ifilt$: options\\passing $C$};
\node[box,right=of i] (r) {\textbf{Choice}\\$R$: ranking\\$p$: final pick\\under $P$};
\foreach \a/\b in {q/f,f/c,c/i,i/r} \draw[arrow] (\a) -- (\b);
\end{tikzpicture}
\caption{A structured reasoning process for product advice.
The request and background knowledge yield facts and beliefs
that inform constraints and preferences.
Constraints filter the candidate pool; preferences guide ranking to
select a recommendation. This flow helps locate shortcomings in LLM advice.}
\label{fig:framework}
\end{figure*}
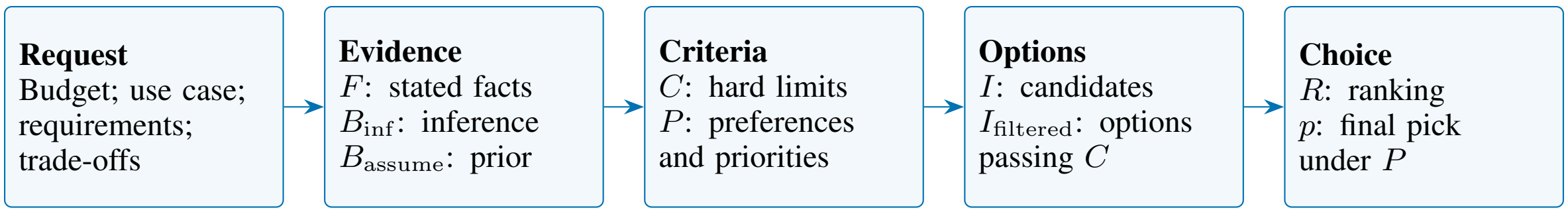

We diagnose advice by separating its stages, from interpreting the
request to selecting a product (Figure~\ref{fig:framework}). Explicitly
listed products form the \emph{candidate pool}. Changing ``SUV or minivan''
to ``SUV only'' calls for a different correction from choosing a five-seat
car after correctly recording the six-seat requirement. Our framework
defines 19 possible failure modes and the evidence needed to assess each.
It examines the request and the written response, so it does not require
access to a model's internal computation.

Assessing quality requires adequate product evidence. Human responses
to the same request provide a narrower comparison: we call a product
name taken from such a response and used as a comparison target a
\emph{reference}. If a reply names a Toyota Sienna, a model's Honda
Odyssey is a mismatch; judging its suitability still requires vehicle
facts and the user's needs. Community references are neither exhaustive
nor guaranteed good recommendations. Advice benchmarks address multiple perspectives
through community preferences or expert
rubrics~\citep{kim2024advisorqa,kumar2025compo,fan2026shoppingreasoning},
and prior work distinguishes matching a recorded product from being
useful to a user~\citep{wang2023rethinking}.

We analyze saved recommendations from seven models. Listing options
tests whether a reference enters the candidate pool; choosing from a
supplied list tests selection with those options available. We compare
choices with and without the user's request and test selection on a
separate shopping-search dataset with relevance labels. Naming a
reference does not ensure selecting it, and measured agreement depends
on answer extraction and list construction. These comparisons locate
behavioral differences; judging quality requires checking whether the
products satisfy the request.

We propose a procedure for assessing the quality and self-consistency
of LLM product advice and locating weaknesses, with reference-based
tests when direct quality judgments are unavailable
(Section~\ref{sec:methodology}). We define a structured representation
of advice and 19 possible failure modes, specifying the evidence
needed to distinguish them (Section~\ref{sec:framework}). We apply
these measurements to saved product recommendations and supplied-choice
experiments, identifying both selection behavior and evaluation
artifacts that affect its interpretation
(Sections~\ref{sec:methodology} and~\ref{sec:coverage-selection}).

\section{Related work}
\label{sec:related}

\paragraph{Advice and reference validity.}
Conversational recommenders elicit preferences~\citep{jannach2021survey}; LLMs support zero-shot
recommendation~\citep{he2023zeroshotcrs}. Evaluation combines
component tests and user judgments~\citep{jannach2023evaluating},
with interactive evaluation exposing limits of recorded-item
matching~\citep{wang2023rethinking}.
AdvisorQA and ComPO use community feedback and
preferences~\citep{kim2024advisorqa,kumar2025compo};
ShoppingReasoningBench and ShoppingComp use
expert-authored tasks and rubrics~\citep{fan2026shoppingreasoning,tou2025shoppingcomp}.
Complementing ShoppingReasoningBench's taxonomy and expert-validated rubric
evaluation, we use coverage and selection stages to examine community-reference
and extraction effects, candidate-construction artifacts, and generation--comparison disagreement.

\paragraph{Constraints, candidates, and ranking.}
Constraint-based recommenders match user requirements to product
properties~\citep{felfernig2015constraint}. Candidate generation and
ranking are established recommendation stages~\citep{covington2016deep},
with retrieval bottlenecks in LLM systems~\citep{dong2026retrievalbottleneck}.
LLM recommendations also exhibit position and popularity
effects~\citep{hou2024zeroshotrankers,lichtenberg2024popularity}.
ESCI supplies external product-search relevance labels~\citep{reddy2022esci}.
We connect these stages to the explicit constraints and evidence
needed to assess a product choice.

\paragraph{Behavioral and factual assessment.}
CheckList tests capabilities beyond aggregate
accuracy~\citep{ribeiro2020checklist}; FActScore checks atomic claims
against reliable sources~\citep{min2023factscore}.
Process supervision evaluates intermediate steps~\citep{lightman2023lets},
but written chains of thought can misrepresent what drives a
response~\citep{lanham2023faithfulness,turpin2023languagemodels}.
We use observable structure to select checks and test interventions,
without treating the response as a trace of internal computation.

\section{Diagnosing advice quality and reference agreement}
\label{sec:methodology}

The diagnostic task takes a user's request and a model's response as
inputs. We want to establish whether the advice meets the user's
needs and identify where it falls short. We first specify the evidence available for judging
quality, then inspect the response's components, and finally test a
suspected weakness by changing the relevant part of the task.

\subsection{Advice quality and reference agreement}
\label{sec:assessment}

For the family-car request, checking the recommended configuration's
seating capacity and price directly tests the firm requirements.
Assessing the running-cost trade-off also requires relevant product
facts and the user's priorities. If the configuration or its price
is unknown, the corresponding requirement cannot yet be judged.
This is the evidence needed for a quality assessment, regardless of
whether anyone else named the same car.

When such judgments are unavailable, a \emph{reference} provides a
narrower comparison. In our community data, it is a product name
extracted from an online reply to the same request; the recorded
names form the \emph{reference set} $S$.
\textbf{Top-reference agreement} measures how often the final pick
matches the highest-weighted entry; \textbf{set-reference agreement}
accepts any entry. For example, if replies name a Toyota Sienna and
a Honda Odyssey, with the Sienna ranked first, choosing either counts
toward set agreement, but only the Sienna counts toward top-reference
agreement.

Reporting both shows how reference choice affects agreement with
the same model outputs. Sparse feedback may not establish a reliable
preference for the top entry, while a single mention does not
establish suitability. Section~\ref{sec:reference} reports both scores
alongside a direct comparison between the model's own recommendation
and a community-listed alternative.

\subsection{Experimental setup}
\label{sec:setup}

\paragraph{Models and data.}
We analyze saved outputs from GPT-5, Claude Opus, Llama-70B, Qwen-80B,
Nemotron-49B, Llama-8B, and Mistral-14B. The Reddit generation corpus
contains 800 car, 800 laptop, and 226 headphone requests from early
2019. The five-domain choice probe additionally uses Stack Exchange Software
 Hardware Recommendations.
The five-domain supplied-choice experiments cover four models. The main
enumeration table covers Llama-70B; supplementary analyses add
six-model coverage and GPT-5.4 generated-pool choice results.
Appendix~\ref{app:coverage} gives
experiment-specific counts and Appendix~\ref{app:repro} gives model
identifiers and prompts. The main text emphasizes car-domain
diagnostics while retaining cross-domain summaries and the ESCI
product-search study. Further results and experimental
details are reported in the corresponding appendices.

\paragraph{Dataset statistics.}
Table~\ref{tab:dataset-statistics} summarizes the requests used
in the experiments. The
Reddit rows describe the shared seven-model generation cohorts; each
model produced three responses per request. The Stack Exchange rows
describe the software and hardware cohorts used in Section~\ref{sec:lift}.
Community replies are counted separately from generated
responses: they include all stored top-level comments on Reddit and answers
on Stack Exchange, even these without recommendations.

\begin{table}[t]
\centering
\small
\setlength{\tabcolsep}{4pt}
\begin{tabular}{@{}lrrrr@{}}
\toprule
Domain & Queries & Names & Replies & Score \\
\midrule
Cars & 800 & 1,681 & 5 & 2 \\
Laptops & 800 & 737 & 4 & 1 \\
Headphones & 226 & 481 & 3 & 1 \\
SE software & 2,445 & 2,349 & 1 & 1 \\
SE hardware & 1,266 & 1,397 & 1 & 1 \\
\bottomrule
\end{tabular}
\caption{\textbf{Dataset statistics.} Names: distinct reference strings;
Replies: median replies/query; Score: median stored reply score.
Full statistics: Appendix~\ref{app:coverage}.}
\label{tab:dataset-statistics}
\end{table}

The median request has five, four, and three stored top-level
replies for cars, laptops, and headphones, respectively, versus one
answer in each Stack Exchange domain. Median reply scores are only
one or two. These counts describe the amount of observed community
feedback.

\paragraph{Community references.}
The Reddit records contain product strings extracted from top-level
comments, weighted by $\max(\mathrm{comment\ score},1)$ per mention.
This process can include comparisons, negative mentions, and malformed
names. We therefore call the raw records \emph{reference mentions}.
Stack Exchange
references use consensus product extraction. Catalog validation checks
product identity; endorsement and suitability require separate evidence.
Appendix~\ref{app:reanalysis-protocol} details the product-matching and
answer-extraction rules, sample aggregation, experimental comparisons,
and uncertainty estimation used in the retrospective reanalysis.

\subsection{Experiments}
\label{sec:results}

We first test whether recommendations satisfy explicit user
constraints, independently of community references. We then test
whether models prefer a supplied community alternative to their own
initial recommendation and examine how reference granularity affects
agreement scores.

\subsubsection{Objective constraint satisfaction}
\label{sec:objective}

The saved free-form car corpus contains responses from all
seven models: three responses to each of 800 shared requests.
The audit covers 149 requests with explicit drivetrain or
transmission requirements, totaling 3,129 responses. The check uses constraints
extracted from the request text shown to the models and a shared
vehicle-fact table, without using the community recommendation as
ground truth. Deterministic rules identify candidate requirements
and exclusions, retaining their source spans for review. Product
facts are reused across recommendations of the same vehicle.

We checked primary recommendations against drivetrain and transmission
requirements using saved US Environmental Protection Agency (EPA)
records, Canadian fuel-consumption records, and manufacturer
specifications. Ambiguous
vehicle identities, configurations, or unsupported facts remain
unknown; unverified catalog attributes are not treated as ground
truth. Shared identity rules account for the stated year, variant,
and purchase market. We reviewed all surfaced candidate conflicts
in their full responses, distinguishing unretracted violations from
disclosed trade-offs and corrected recommendations. Support concerns
only the audited configuration fields; budget and other attributes
remain outside this check. Appendix~\ref{app:reanalysis-protocol}
details the evidence, review, and uncertainty estimates.

\begin{table}[t]
\centering
\small
\begin{tabular}{@{}lrrr@{}}
\toprule
Model & Supported & Conflicts (\%) & Other \\
\midrule
GPT-5 & 80 & 0 (0.0) & 367 \\
Claude Opus & 44 & 0 (0.0) & 403 \\
Llama-70B & 62 & 6 (1.3) & 379 \\
Qwen-80B & 57 & 3 (0.7) & 387 \\
Nemotron-49B & 59 & 3 (0.7) & 385 \\
Llama-8B & 36 & 5 (1.1) & 406 \\
Mistral-14B & 18 & 17 (3.8) & 412 \\
\bottomrule
\end{tabular}
\caption{\textbf{Conservative configuration checks on car advice.}
Each row retains 447 answers (149 requests, three generations).
Supported counts have evidence for the primary recommendation's audited
drivetrain/transmission. Conflicts are counts (\% of 447). Other includes
unresolved and separately classified cases. Conflicts are minimum detected violations;
zeros do not establish error-free advice. Intervals:
Table~\ref{tab:objective-audit}.}
\label{tab:objective}
\end{table}

\paragraph{Results.}
Table~\ref{tab:objective} records 34 confirmed conflicts and 356
supported primary configurations; 2,739 answers remain in the
other/unknown category. Mistral-14B has 17 confirmed conflicts among
447 answers (3.8\%); the other open models have three to six each.
No conflict was confirmed for GPT or Claude under this audit.
A separate 44-query headphone audit finds 795 supported connection
checks, no confirmed conflicts, and 129 unknowns across 924 answers
(Appendix~\ref{app:reanalysis-protocol}).

\paragraph{Discussion.}
Even a narrow specification check identifies concrete failures to
meet explicit requirements. However, exact product identity and
market-specific evidence limit how many answers can be assessed.
The supported and conflict cases cover only 390 of 3,129 car answers,
with no query/sample position assessable for all seven models.
The counts therefore establish minimum detected violations, while
a comparative compliance ranking requires broader evidence coverage.
Constraint compliance does not capture every preference or
trade-off. We next examine what community references contribute to
the evaluation and what their limitations are.

\subsubsection{Comparing to the reference}
\label{sec:reference}
\label{sec:preference}

A recommendation can differ from a community reference without
being worse. We ask whether the same model, shown its own recommendation
and a community-listed alternative, prefers the alternative for the original
request. This measures agreement between initial generation and comparative
judgment under different prompts.

\paragraph{Source-blind comparison.}
For each of 150 shared car requests, we fix a community-listed
alternative before generating three initial recommendations per
model. Whenever a valid recommendation differs from that alternative,
we ask the same model to compare the two in a fresh context, using
the full request but omitting the initial explanation and both
options' sources. We present the pair in both orders, given reported
position biases in LLM judging~\citep{zheng2023judging}, allowing ties
or abstention rather than forcing a choice. Appendix~\ref{app:preference} provides the
alternative-selection rule, prompts, and response formats.

\paragraph{Results.}
Table~\ref{tab:preference} shows that the community-listed
alternative wins in both orders for 1.9--6.5\% of eligible initial
recommendations across all six models with completed runs, after the
product-identity sensitivity check. GPT-5.4 and Llama-70B each have
26 such trials (6.5\% and 6.4\%), while Claude, Qwen, and Nemotron
have rates of 4.1--4.5\%. Ministral-14B was unavailable on the tested
endpoints and is not included. Temperature-zero initial generation yields
5.3\% for GPT and 3.1\% for Claude (Appendix~\ref{app:temperature-zero}).

A low community-preference rate can coexist with unstable
choices. Llama-8B consistently favors its original recommendation in
only 59.1\% of eligible trials, compared with 78.5--85.8\% for the
other models. In 94/367 trials (25.6\%), it instead chooses the
first-listed option in both presentations, changing its product
choice when the order changes. Its low rate of community
preference (1.9\%) thus does not establish stronger initial
advice.

\begin{table}[t]
\centering
\small
\begin{tabular}{@{}lrr@{}}
\toprule
Model & Count & Rate [95\% CI] \\
\midrule
GPT-5.4 & 26/402 & 6.5 [3.0, 10.3] \\
Claude Opus 4.8 & 16/392 & 4.1 [1.5, 7.3] \\
Llama-70B & 26/405 & 6.4 [2.9, 10.6] \\
Qwen-80B & 18/400 & 4.5 [2.2, 7.4] \\
Nemotron-49B & 13/317 & 4.1 [1.2, 7.7] \\
Llama-8B & 7/367 & 1.9 [0.5, 3.6] \\
\bottomrule
\end{tabular}
\caption{\textbf{Preference for the community alternative.} Counts:
community choices in both source-blind orders / eligible trials,
from three initial generations on 150 shared car requests per model.
Rates are percentages with 95\% query-cluster bootstrap intervals.
Eligibility and identity exclusions: Appendix~\ref{app:preference}.}
\label{tab:preference}
\end{table}

A restricted laptop replication yields 44/400 both-order preferences
for GPT-5.4 (11.0\%) and 41/403 for Opus 4.8 (10.2\%) after the
identity sensitivity check (Appendix~\ref{app:laptop-preference}).
Its references are community mentions concentrated in MSI and Dell;
cohort differences preclude a controlled comparison with cars.
The rates describe comparative preferences on each model's eligible
recommendations. Assessing advice quality requires judging the products
against the user's needs.

\paragraph{Agreement with stored references.}
As a separate descriptive comparison, Table~\ref{tab:agreement}
reports how often the seven models' saved recommendations match the
top reference or any name in the reference set, as defined in
Section~\ref{sec:assessment}. With outputs and matcher fixed, accepting
any reference entry raises agreement throughout, but all-entry
agreement remains only 14.7--27.4\% across models and domains.
Several mechanisms could contribute: model failures
(Section~\ref{sec:framework}), community biases toward popular or
familiar products and favored brands, and different weightings of
the user's preferences. For example, a request for a ``large and cheap''
car does not specify the relative importance of space and price.
The community may favor a roomier, costlier car, while the model
favors a smaller, cheaper one, even when both satisfy the hard
constraints. This trade-off mismatch need not indicate an error by
either. Incomplete references and extraction errors can
lower measured agreement. The scores alone cannot distinguish these
explanations. Appendix~\ref{app:agreement} adds a software extraction-scope sensitivity.

\begin{table}[t]
\centering
\small
\setlength{\tabcolsep}{3pt}
\begin{tabular}{@{}lrrr@{}}
\toprule
Model & Cars & Laptops & Headphones \\
\midrule
GPT-5 & 6.8 / 24.8 & 10.3 / 26.0 & 6.6 / 22.6 \\
Claude Opus & 6.5 / 24.0 & 9.6 / 24.8 & 7.2 / 24.6 \\
Llama-70B & 6.0 / 24.5 & 10.0 / 27.4 & 8.3 / 21.5 \\
Qwen-80B & 5.2 / 23.2 & 5.0 / 19.0 & 5.5 / 17.1 \\
Nemotron-49B & 6.0 / 22.0 & 8.6 / 23.3 & 4.9 / 17.7 \\
Llama-8B & 5.6 / 21.3 & 9.0 / 24.0 & 5.2 / 16.1 \\
Mistral-14B & 5.5 / 20.8 & 5.1 / 18.4 & 3.7 / 14.7 \\
\midrule
Queries & 800 & 800 & 226 \\
\bottomrule
\end{tabular}
\caption{\textbf{Reference granularity changes agreement.} Top-entry /
all-entry mention agreement (\%), using identical outputs, the same
matcher, and all three samples. Missing picks remain in the denominator.
Query-bootstrap intervals are in Appendix~\ref{app:agreement}.}
\label{tab:agreement}
\end{table}

\section{A structured framework for advice}
\label{sec:framework}

\subsection{Evidence, criteria, and decisions}

We first distinguish the kinds of information needed to assess a
recommendation. In the family-car example, six seats are required,
lower running costs are preferred, and an unrequested assumption
about driving in snow is neither. Keeping these roles separate makes
it possible to ask where an unsupported restriction entered the
answer. A statement supplied by the user, an assumption added by the
model, and a product choice must be checked against different evidence.

We represent a response to request $q$ by
$\mathcal{A}(q)=(F,B,C,P,I,\Ifilt,R)$.
$F$ contains facts explicitly stated or directly supported by the
request. $B$ contains additional beliefs, separating inferences
$\Binferred$ with evidence in $F$ from assumptions $\Bassumed$
introduced as background priors. $C$ contains hard constraints;
$P$ contains soft preferences and their stated priorities.
$I$ is an explicit candidate set, $\Ifilt$ the subset retained after
applying $C$, and $R$ a ranking or recommendation over that subset.
The final pick is $p(q)$.

 This decomposition describes the reasoning tasks involved in giving
product advice: understand the request, identify relevant products,
apply the requirements, and weigh the alternatives. Making
these steps explicit lets us assess different capabilities separately.
For example, a model may preserve the user's requirements yet fail to
name a suitable product, or name one but reject it when choosing.
This emphasis on explicit intermediate structure is related to
chain-of-thought prompting~\citep{wei2022cot} and decomposed
prompting~\citep{khot2023decomposed}, which organize problem solving
into reasoning steps and modular subtasks, respectively.

 The distinction between inference and assumption prevents a common
logical substitution. ``Winters are harsh'' supports asking about
weather-related needs; it does not establish that all-wheel drive is a
hard requirement. Likewise, ``SUV or minivan'' should not silently
become ``SUV only.'' These are interpretable failures even when the
request admits several final products. The framework makes
their evidential requirements explicit.
The representation is limited to observable content: the model may
consider additional candidates, preferences, or priors without stating them.
Prompts requesting structured output can expose these fields, but
they may change the recommendation.

\subsection{Failure taxonomy}
\label{sec:taxonomy}

The taxonomy contains 19 modes in six categories: fact extraction,
belief attribution, constraint and preference formation, candidate
construction, filtering, and ranking. Appendix~\ref{app:taxonomy}
gives the complete inventory and the evidence each diagnosis requires.
For example,
filtering a satisfying option out and retaining a violating option
are distinct from failing to generate either option in the first place.

We implement a \emph{prototype} set of automated checks, which we call
verifiers.  The selected
implementation combines 15 non-generative heuristics with four
generative judge checks. The
non-generative checks include lexical coverage, citation-tag patterns,
constraint patterns, and catalog comparisons. When the evidence
required for a check is missing, its outcome is unassessable rather
than a model error.

\paragraph{Worked example: a wrong constraint value (3b).}
Suppose a request requires a vehicle for at least six passengers under
\$20,000. The response records both requirements correctly in $F$,
but changes the minimum capacity to five in $C$. This is failure 3b:
a correctly extracted requirement changes during constraint formation,
before candidate generation or filtering. The prototype verifier
flags this repairable mistake without deciding which car is best.
Appendix~\ref{app:implementation} gives the numerical check, its
reproduced scores, and its limitations.

  The experiments below distinguish the behavior of these checks from
the effects of a prompt intervention on the final recommendation.
Root-check flags identify responses for inspection; product matches
measure agreement with the recorded references. Estimating the
prevalence of all 19 failures would additionally require validating
the checks against independently assessed responses.

\subsection{Dependency-guided diagnosis}
\label{sec:dependency}

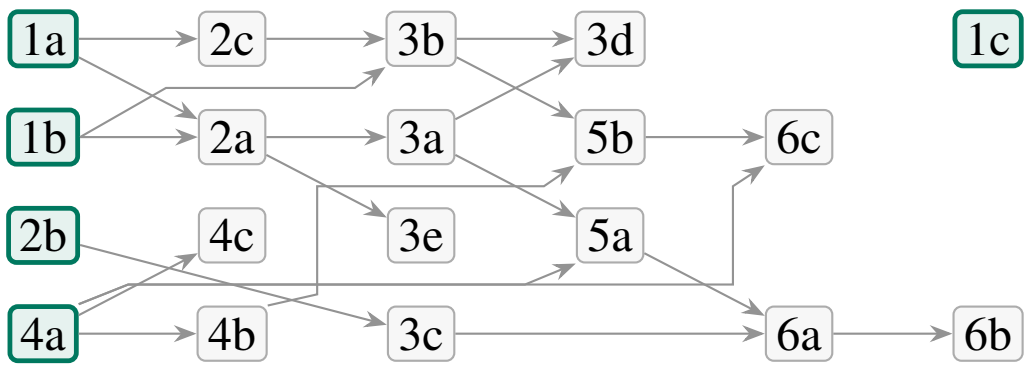
\begin{figure}[t]
\centering
\begin{tikzpicture}[
  font=\small, x=12.5mm, y=6.5mm,
  fm/.style={draw=gray!65,rounded corners=1.5pt,inner sep=2pt,fill=gray!6},
  root/.style={fm,draw=leaf,line width=0.9pt,fill=leaf!10},
  dep/.style={-{Stealth[length=1.5mm]},gray!85}]
\node[root] (1a) at (0,3) {1a}; \node[root] (1b) at (0,2) {1b};
\node[root] (2b) at (0,1) {2b}; \node[root] (4a) at (0,0) {4a};
\node[root] (1c) at (5,3) {1c};
\node[fm] (2c) at (1,3) {2c}; \node[fm] (2a) at (1,2) {2a};
\node[fm] (4c) at (1,1) {4c}; \node[fm] (4b) at (1,0) {4b};
\node[fm] (3b) at (2,3) {3b}; \node[fm] (3a) at (2,2) {3a};
\node[fm] (3e) at (2,1) {3e}; \node[fm] (3c) at (2,0) {3c};
\node[fm] (3d) at (3,3) {3d}; \node[fm] (5b) at (3,2) {5b};
\node[fm] (5a) at (3,1) {5a};
\node[fm] (6c) at (4,2) {6c}; \node[fm] (6a) at (4,0) {6a};
\node[fm] (6b) at (5,0) {6b};
\foreach \a/\b in {1a/2a,1b/2a,1a/2c,2a/3a,2c/3b,2b/3c,
  3a/3d,3b/3d,2a/3e,4a/4b,4a/4c,3a/5a,3b/5b,
  5a/6a,5b/6c,6a/6b}
  \draw[dep] (\a) -- (\b);
\draw[dep] (1b.east) -- (0.65,2.5) -- (1.65,2.5) -- (3b.south west);
\draw[dep] (4a.north east) -- (0.45,0.5) -- (2.55,0.5) -- (5a.south west);
\draw[dep] (4b.north east) -- (1.45,0.4) -- (1.45,1.5) -- (2.65,1.5) -- (5b.south west);
\draw[dep] (4a.north east) -- (0.45,0.5) -- (3.65,0.5) -- (3.65,1.5) -- (6c.south west);
\draw[dep] (3c.east) -- (6a.west);
\end{tikzpicture}
\caption{Proposed diagnostic dependency graph for the
19 failure modes. The five highlighted root nodes are entry checks;
arrows indicate downstream checks to revisit after an intervention.
Codes follow Appendix~\ref{app:taxonomy}.}
\label{fig:dag}
\end{figure}

An upstream error can change the evidence used by later
checks: a misrecorded seating requirement can affect both the derived
constraint and the filtering decision. Figure~\ref{fig:dag} organizes
the taxonomy into a proposed diagnostic ordering. Its five root
checks concern missing facts (1a), invented facts (1b), changed
valence or attribution (1c), priors presented as user-derived (2b),
and inadequate candidate coverage (4a). After addressing a flagged
issue, we revisit the downstream checks using the updated response.
The arrows specify this inspection order; a check can be assessed directly
whenever its required evidence is available.

\subsection{Experiments}
\label{sec:root-experiments}
\label{sec:fact-anchoring}

\textbf{Root-check profiles.} We revisit the archived root-check experiment on structured
car advice. Table~\ref{tab:roots} uses the same 100 requests across
all seven models and the first saved generation for each request.
The query-side checks use the title and the 1,800-character body
available to the original generation prompt. We report each
verifier's explicit firing flag, since the checks use different
score thresholds. Appendix~\ref{app:implementation} gives the
operational rules and the reconciliation with the archived scores.
The profiles vary by check. The assumption-tag check (2b)
never fires for Claude or Llama-70B, but fires on 41 GPT and
35 Mistral responses. The candidate-count check (4a) flags two
Claude responses and 42 Mistral responses. By contrast, lexical
fact checks fire frequently across models: 45--70 responses for
1a and 59--99 for 1b. Paraphrases can trigger these lexical checks,
while correct tagging alone does not establish that a belief is
justified. The profiles identify different responses and fields to
inspect. Appendix~\ref{app:implementation} compares two checks with
exploratory assistant annotations.

\begin{table}[t]
\centering
\small
\begin{tabular}{@{}lrrrrr@{}}
\toprule
Model & 1a & 1b & 1c & 2b & 4a \\
\midrule
GPT-5 & 45 & 73 & 7 & 41 & 5 \\
Claude Opus & 51 & 85 & 3 & 0 & 2 \\
Llama-70B & 70 & 59 & 3 & 0 & 21 \\
Qwen-80B & 63 & 88 & 4 & 23 & 39 \\
Nemotron-49B & 67 & 96 & 2 & 19 & 25 \\
Llama-8B & 70 & 71 & 2 & 14 & 20 \\
Mistral-14B & 57 & 99 & 3 & 35 & 42 \\
\bottomrule
\end{tabular}
\caption{\textbf{Root-check flags on structured car advice.}
Counts out of 100 shared requests per model, one generation each.
Checks measure lexical fact coverage (1a), token/number
support (1b), polarity patterns (1c), assumption tags (2b), and
candidate-count shortfalls (4a).}
\label{tab:roots}
\end{table}

\textbf{Fact anchoring.} Does supplying the fact list improve the final recommendation?
An archived Llama-70B experiment prefills $F$ with request sentences
on 400 paired car requests, with three generations per condition.
Reference agreement is 22.9\% without anchoring and 22.0\% with it
($-0.9$ points; 95\% paired query-bootstrap interval $[-2.8,+0.8]$).
An input-length-matched sensitivity also finds no improvement.
The model can rewrite the supplied list, so the targeted representation
and final answer both need checking. Appendix~\ref{app:implementation}
gives the protocol and controls.
The root checks reveal diagnostic
profiles across models: Claude has fewer candidate-count flags than
Llama-70B but more token-support flags. This variation motivates
component-specific follow-up.
Fact anchoring also fails to improve reference agreement, highlighting
the need to verify both the targeted representation and the downstream
effect of an intervention. These findings concern car advice;
validated failure rates and differences between domains remain open.

\section{Candidate coverage and selection}
\label{sec:coverage-selection}

Coverage and selection capture two critical failure points: omitting
all suitable products from the pool, or failing to choose one when
available. Either can leave the user without suitable advice despite
correct reasoning elsewhere. We measure these stages against recorded
reference products.

\textbf{Candidate coverage} is the fraction of requests whose explicit
pool contains at least one reference, addressing failure 4a
(inadequate candidate coverage). \textbf{Conditional selection}
measures how often a reference is chosen when the pool contains one.
For a reference such as the Honda Odyssey, we test its inclusion if
absent and inspect filtering (5a--5b) or ranking (6a) if present but
unchosen. This follows the candidate-generation and ranking
distinction~\citep{dong2026retrievalbottleneck}. Diagnosing a failure
still requires evidence about product suitability, constraints, and
preferences.

For choices restricted to the pool, reference agreement requires both
coverage and selection. If choices can name products outside the pool,
those matches must be counted separately.

\subsection{Candidate coverage}
\label{sec:enumeration}

\paragraph{Question and test.}
Does the model generate reference products, and does coverage
vary by domain? We supply the request without community product names
and ask for three lists of 20 plausible products. We merge names in
first-mention order and measure whether a recorded reference appears
among the first 10 or 40 candidates. This enumeration experiment
includes Llama-70B on 171 car, 176 laptop, and 49 headphone requests.
Appendix~\ref{app:enum} gives the protocol and extends coverage to six
models on shared subsets of 160, 164, and 49 requests, respectively.

\begin{table}[t]
\centering
\small
\setlength{\tabcolsep}{3pt}
\begin{tabular}{@{}lrrrr@{}}
\toprule
& & \multicolumn{2}{c}{Coverage (\%)} & Selection \\
\cmidrule(lr){3-4}
Domain & $n$ & $K=10$ & $K=40$ & \% ($n_c$) \\
\midrule
Cars & 171 & 59.6 & 76.6 & 26.5 (131) \\
Laptops & 176 & 85.8 & 95.5 & 47.0 (168) \\
Headphones & 49 & 49.0 & 57.1 & 39.3 (28) \\
\bottomrule
\end{tabular}
\caption{\textbf{Llama-70B coverage and selection.} Coverage counts
reference-containing pools; selection averages three choices on
$n_c$ covered queries at $K=40$ (invalid: zero).
Extensions and controls: Appendix~\ref{app:enum}.}
\label{tab:enum}
\end{table}

\paragraph{Results.}
At 40 candidates, reference coverage is 95.5\% for laptops,
76.6\% for cars, and 57.1\% for headphones (Table~\ref{tab:enum}).
Thus, even after three enumeration calls, a recorded reference is
absent from 42.9\% of headphone pools, compared with 4.5\% of laptop
pools. The six-model appendix analysis also shows higher coverage for laptops
than headphones in every model, with different point-estimate orderings
across domains. The appendix includes catalog- and source-reference sensitivities.
Extending a list necessarily preserves or increases coverage; whether
the model then selects a reference is a separate question.

\paragraph{Discussion.}
Why might a product be missing? One possibility is weak or
absent product knowledge; another is that the model knows the product
but the request does not make it sufficiently likely to be generated.
A finite list also excludes lower-ranked products,
and the model may judge a community reference unsuitable. Coverage
therefore measures what is brought into the explicit pool, not
everything the model knows.

An archived open-model likelihood-lift experiment scores reference
names with and without the request. In 156 matched Llama-70B car
requests, 28/69 omitted target references (40.6\%) receive positive
lift: context can raise a product's likelihood without bringing it
into the pool. Prompt and serving differences limit this association.
Appendix~\ref{app:enum} reports the three-model, five-domain results
and controls.

\subsection{Selection}
\label{sec:lift}

\paragraph{Question and tests.}
When a reference is available, how often does the model select
it, and how does this vary across models and domains? We first ask
Llama-70B to choose three times from each generated pool, measuring
agreement only where a reference is present. To compare four models
across five advice domains, we also supply one recognizable community
reference and four popular nonmatching distractors. We present the
same shuffled options with a ``best in general'' instruction and with
the request and a ``best for this user'' instruction. This controls
reference availability and tests the effect of the request-specific
prompt; both the request and instruction change.
Appendix~\ref{app:lift} reports both conditions and paired intervals.

\paragraph{Results across models and domains.}
Within its own reference-containing pools, Llama-70B selects
a reference on 26.5\% of car, 47.0\% of laptop, and 39.3\% of
headphone choices (Table~\ref{tab:enum}). GPT-5.4
selects a reference in 30.9\% of covered car choices and 24.0\% of covered
headphone choices (Appendix~\ref{app:enum}); these rates condition on
different model-generated pools. Laptop coverage is therefore
near-complete, yet selection agrees on fewer than half of eligible
choices. In the supplied five-option test, the request-specific
prompt increases agreement in all 20 model--domain cells.
Agreement is 46.0--62.5\% across the Reddit domains,
compared with 91.3--96.7\% on Stack Exchange
(Appendix~\ref{app:lift}, Table~\ref{tab:lift}).

\paragraph{Interpreting the domain differences.}
The supplied lists admit a shortcut: the inserted reference is
often less popular than the four distractors. Choosing the least-popular
option in reconstructed pools, without reading the request, exceeds
every model's point estimate in each domain (Table~\ref{tab:lift}).
This control diagnoses the benchmark using construction metadata
unavailable in the model prompts.
The observed domain differences therefore cannot establish that models
handle user preferences better in one domain than another.
In a balanced-pair test, most of the real-versus-shuffled request gain
comes from transitions between non-selection and choosing the reference
(Appendix~\ref{app:lift}).

\paragraph{External relevance check.}
\label{sec:esci}
To test selection against assigned relevance labels, we use
ESCI product-search queries~\citep{reddy2022esci}. Each prompt
contains one Exact-labeled product and three Irrelevant-labeled
products, shuffled without labels, with titles truncated to 120
characters. On a custom sample of 500 US queries, Claude and GPT-5
select the Exact product in 73.6\% and 70.2\% of cases; Llama-70B
and Llama-8B score 67.4\% and 58.6\% (uniform choice: 25\%).
Thus even the two closed models miss the product on
26.4\% and 29.8\% of queries. Appendix~\ref{app:esci} gives sampling
details, uncertainty estimates, and the Qwen-80B result
(67.8\%). Truncated titles, label noise, and invalid outputs can
contribute to misses, and the saved hit flags do not distinguish them.

\paragraph{Conclusion.}
High reference coverage does not ensure reference selection.
Selection scores vary across models and domains, but the community
test's shortcut limits their interpretation; ESCI provides a separate
relevance-label check.
The free-form versus enumeration comparison and its extraction audit
are retained in Appendix~\ref{app:enum}, since their different answer
interfaces prevent attributing the score difference to better advice.

\section{Discussion}
\label{sec:discussion}
\label{sec:conclusion}

We investigated product advice from LLMs, distinguishing objective
quality from community agreement and decomposing advice into 19
possible failure points. Coverage and selection deserve separate
evaluation: either can prevent a suitable product from reaching the
final recommendation.

Even frontier models sometimes prefer an alternative to their initial
advice: GPT-5.4 and Claude Opus 4.8 favor the community-listed product
in some comparisons in both source-blind orders
(Table~\ref{tab:preference}). This reveals a mismatch between
generation and comparative choice. Root checks flag omitted user
facts and unsupported additions, with different profiles across models
(Table~\ref{tab:roots}). These flags need inspection because lexical
checks can flag valid paraphrases. Supplying a fact list did not
improve reference agreement, so repairs must be checked at both the
targeted stage and the final answer.
Reference coverage and selection vary across domains, with high coverage
sometimes accompanied by low selection agreement (Table~\ref{tab:enum}).

The next step is to improve advice through targeted interventions:
catalog-backed generation for missing candidates, or better filtering
and ranking when suitable candidates are rejected. Validating the
diagnostic checks and testing models on verified suitable candidates
can help choose the intervention. Its effect should be assessed
against verified constraints and independent suitability judgments,
using comparable answer interfaces and inference budgets.

\section*{Limitations}
\label{sec:limitations}

\paragraph{Reference and extraction validity.}
Raw community mentions are neither verified endorsements nor an
exhaustive set of acceptable answers. Catalog-based extraction misses
unlisted products and can select a mentioned alternative instead of
the intended recommendation. A prior automated endorsement check on
250 catalog-validated entries does not validate every raw entry used
here. Independent human
validation remains necessary for claims about acceptable disagreement.

\paragraph{Observable structure and verifier scope.}
The taxonomy describes the explicit response.
The prototype combines heuristics and generative judges and has not
been validated as a 19-way detector on representative human-labeled
responses. Format restrictions can affect model
performance~\citep{tam2024speakfreely}; structured prompting can
therefore alter the output being examined.
Accordingly, no claim about the dominant failure mode is based on its
uncalibrated firing counts.

\paragraph{Temporal and population scope.}
The main Reddit requests date to 2019, while the models and some
catalog entries are newer. The matching reanalysis does not enforce
historical availability, so it cannot certify advice that was feasible
when the request was posted. The corpora are English-language,
platform-specific, and concentrated in a US-oriented product space.
Reference extraction also selects for recognizable products; neither
the queries nor the chosen models are a random sample of all advice.

\paragraph{Experimental controls and saved outputs.}
Enumeration uses more calls than direct answering and changes the
prompt. Its asymmetric answer extraction prevents treating the laptop
score difference as an established recommendation gain. Query lift
uses generic popular distractors, admits a popularity-rank shortcut,
and changes the task instruction. ESCI uses truncated titles, a custom surface
filter, and no same-pool lexical baseline. Its archived hit flags, and
the lift records, cannot recover raw parsing failures. These constraints
limit causal interpretation while leaving the reported protocol-level
measurements reproducible.

The source-blind comparison preserves the request and counterbalances
option order, but changes the task from generating an answer to comparing
supplied names. Reference names are often less specific than initial
recommendations, so year, trim, and configuration information need not
be balanced. The identity check removes explicit duplicate or ambiguous
pairs. Product facts, historical availability, community endorsement,
and the models' explanations require independent validation before a
preference reversal can be interpreted as a correction of bad advice.
The identity exclusions are post-hoc
and the frozen-cohort result is reported alongside them.

\paragraph{Uncertainty and reproducibility.}
Query-level intervals do not include generation-seed variance, and
multiple exploratory comparisons do not establish a complete model
ranking. Model identifiers are recorded, but full served
version and timestamp provenance is incomplete for some historical
runs. The manuscript combines an offline reanalysis of historical
outputs with the separately archived full-request comparison run.
Neither supplies independent human judgments of recommendation quality.

\section*{Ethical considerations}

\paragraph{Public user text.}
The study uses public Reddit and Stack Exchange requests and responses.
The working
data contain raw text, which may include identifying details even
without dedicated username fields.

\paragraph{Consumer and exposure risks.}
Community popularity can favor well-known products without satisfying an
individual's budget or needs, and optimizing it can reinforce exposure
imbalances.

\paragraph{Compute and assistance.}
The reported model experiments use inference; enumeration requires generation
calls as well as choices. The full-request car comparison contains 4,895
responses for GPT and Claude (September 27, 2026), including reconsideration
controls, and 4,834 responses for four open models (October 10). GPT-5.4 car
and headphone enumeration/choice and Qwen ESCI use 1,820 responses collected
on October 10. Laptop source-blind comparisons, including parser recovery,
use 1,254 GPT responses (812,486 tokens) and 1,269 Opus responses
(1,317,156 tokens). Balanced-pair selection uses 1,128 GPT/Llama responses
(463,570 tokens), 564 Opus responses (371,275 tokens; October 11), and
564 Qwen responses (238,619 tokens; October 12). GPT-5.4 laptop enumeration
and choice uses 1,056 responses (798,996 tokens; October 11), and Opus headphone
enumeration and choice uses 294 responses (218,820 tokens; October 12).
The temperature-zero initial-generation check uses 826 responses
(517,099 reported tokens; October 13). Objective and reference audits use
saved answers, and the paired transition analysis uses the 2,256 balanced-pair
responses. These analyses require no further evaluated-model inference.
Generative assistants supported code, analysis, and manuscript revision.
The authors remain responsible for checking the final experimental claims,
citations, data handling, and submission disclosures.

\begingroup
\apptocmd{\thebibliography}{\fontsize{10pt}{12pt}\selectfont}{}{\errmessage{Bibliography font patch failed}}
\bibliography{references}
\endgroup

\appendix

 \section{Experiment coverage and data provenance}
\label{app:coverage}

Table~\ref{tab:dataset-statistics-full} preserves the full dataset
statistics summarized in Table~\ref{tab:dataset-statistics}.
Table~\ref{tab:coverage} maps each experiment to its domains, models,
and query counts.

\begin{table*}[t]
\centering
\small
\begin{tabular}{@{}lrrrrrr@{}}
\toprule
Domain & Requests & \shortstack{Distinct\\ref. names} & \shortstack{Replies\\(total)} & \shortstack{Replies\\per query} & \shortstack{Ref. names\\per query} & \shortstack{Score\\per reply} \\
\midrule
Cars & 800 & 1,681 & 5,195 & 5 [3, 7] & 5 [4, 8] & 2 [1, 3] \\
Laptops & 800 & 737 & 3,092 & 4 [3, 5] & 4 [3, 5] & 1 [1, 2] \\
Headphones & 226 & 481 & 968 & 3 [2, 6] & 3 [3, 5] & 1 [1, 2] \\
SE software & 2,445 & 2,349 & 4,663 & 1 [1, 2] & 1 [1, 2] & 1 [1, 2] \\
SE hardware & 1,266 & 1,397 & 2,053 & 1 [1, 2] & 1 [1, 1] & 1 [1, 3] \\
\bottomrule
\end{tabular}
\caption{\textbf{Full dataset statistics.} The final three columns give medians
[25th, 75th percentiles]. Reference names are case- and
whitespace-normalized; aliases and extraction errors remain. Reply scores
are stored Reddit or original Stack Exchange scores, excluding accepted-answer
bonuses; they are not separate upvote counts.}
\label{tab:dataset-statistics-full}
\end{table*}

\begin{table*}[t]
\centering
\small
\begin{tabular}{@{}p{4.2cm}p{3.0cm}p{3.2cm}p{3.4cm}@{}}
\toprule
Experiment & Domains & Models & Queries per cell \\
\midrule
Reference-granularity reanalysis & Cars, laptops, headphones & All seven & 800 / 800 / 226 \\
Objective configuration audit & Cars & All seven & 149 shared; 3 saved generations each \\
Objective connection audit & Headphones & All seven & 44 shared; 3 saved generations each \\
Matched enumeration and choice & Cars, laptops, headphones & Llama-70B & 171 / 176 / 49 \\
Recovered paired coverage & Cars, laptops, headphones & Six; no Mistral & 160 / 164 / 49 \\
Generated-pool choice & Cars; laptops; headphones & GPT-5.4 & 171 / 176 / 49; plus 171 archived car queries \\
Generated-pool choice & Headphones & Opus 4.8 & 49; same requests and protocol \\
Audited hardware enumeration and choice & SE hardware & Llama-70B & 234 archived; 194 with specific-product references \\
Teacher-forced likelihood lift & All five & Llama-8B, Llama-70B, Qwen-80B & 1,292 / 902 / 198 / 2,445 / 1,266 \\
Likelihood lift joined to pool inclusion & Cars & Llama-70B artifacts & 156 matched; 577 archive sensitivity \\
Query-conditioned supplied choice & All five & GPT-5, Claude, two Llamas & 1,292 / 902 / 198 / 2,445 / 1,266 \\
External-reference choice & ESCI US custom sample & GPT-5, Claude, two Llamas, Qwen & 500 \\
Source-blind own-answer comparison & Cars, full requests & Six; Ministral unavailable & 150 shared; 3 initial generations each \\
Temperature-zero initial generation & Cars, full requests & GPT-5.4, Opus 4.8 & Same 150; 1 initial generation each \\
Source-blind laptop comparison & Restricted laptop cohort & GPT-5.4, Opus 4.8 & 150; 3 initial generations each \\
Root-check profiles & Cars, structured advice & All seven & 100 shared; first generation \\
Assistant root-check pilot & Cars, structured advice & All seven & 20 shared; first generation \\
Software extraction sensitivity & SE software & All seven & 400 shared; 3 saved generations each \\
Headphone source/identity sensitivity & Headphones & Six coverage; three choice runs & 49 fixed requests \\
Balanced-pair selection & Cars & GPT-5.4, Opus, Llama-70B, Qwen-80B & 94 shared; 3 conditions, 2 orders \\
Fact-anchored prompt & Cars, structured advice & Llama-70B & 400 paired; 361 input-length matched \\
Raw comment-support count & Cars reference corpus & No model generation & 1,465 \\
\bottomrule
\end{tabular}
\caption{\textbf{Coverage of the results in this manuscript.}
Corpus size and evaluated sample size are distinct. The uniform-slot
baseline uses the same enumeration pools as its chooser comparison.}
\label{tab:coverage}
\end{table*}

Reddit requests come from \texttt{r/whatcarshouldIbuy},
\texttt{r/SuggestALaptop}, and \texttt{r/HeadphoneAdvice}, with the
main generation corpus drawn from January--April 2019. Eligibility
required a personal recommendation request and at least three weighted
raw reference entries. Reference-support and lift analyses use larger
available reference corpora where specified. Stack Exchange data were drawn from 2024 dumps.
The documented extraction funnel attempted 3,000 software and 2,352
hardware questions and retained 2,445 and 1,266 with a consensus
reference. Retention is conditional on successful extraction, and the
extractor models overlap with evaluated models.

\section{Reproduction, prompts, and model identifiers}
\label{app:repro}

\paragraph{Evaluation procedure.}
The experiment-specific appendices specify the cohorts, prompts, scoring rules,
aggregation, and uncertainty estimates. Agreement and enumeration analyses use
saved generations and choices, with query-level resampling that keeps each
request's repeated samples together. The model/domain extensions follow the
corresponding protocols, with any changes and remaining validation limits stated
alongside their results.

\paragraph{Models.}
The model labels correspond to the following requested serving identifiers:
\begin{itemize}[leftmargin=1.4em,itemsep=1pt]
\item Llama-8B: \path{meta/llama-3.1-8b-instruct}.
\item Llama-70B: \path{meta/llama-3.3-70b-instruct}.
\item Qwen-80B: \path{qwen/qwen3-next-80b-a3b-instruct}.
\item Nemotron-49B: \path{nvidia/llama-3.3-nemotron-super-49b-v1}.
\item Mistral-14B: \path{mistralai/ministral-14b-instruct-2512}.
\item GPT-5: \path{openai/gpt-5.4}.
\item Claude Opus: \path{anthropic/claude-opus-4-8}.
\end{itemize}
These identifiers document the requested endpoints. Generation timestamps and
endpoint-version histories are not complete for all saved records.

\paragraph{Free-form generation.}
Free-form generation uses a single user message with the template:
\begin{quote}\small
You are a [product] recommendation advisor. Read the user's request
and give one practical recommendation. Start with your top
recommendation as a specific [product] name, then briefly explain why
it fits. You may mention alternatives after the top recommendation.
User's request: Title: [title] [body]
\end{quote}
It requests three samples at temperature 0.7, nominal seeds 42--44,
with body truncation at 1,800 characters. Output limits are 4,096 tokens for the two frontier models and 1,100 for the open
models. The reanalysis uses the raw saved responses.

\paragraph{Enumeration and selection.}
Enumeration and selection each receive one user message. In both, [body] is
truncated to 1,500 characters and [product] denotes the domain's product category.
The enumeration prompt is:
\begin{quote}\small
A user asks for a [product] recommendation:\\
Title: [title]\\{}
[body]\\
List 20 DISTINCT specific [product] options that could plausibly fit this
request -- cast a wide net, include less obvious ones, and give exact product
names. One per line, name only, no commentary.
\end{quote}
The chooser receives:
\begin{quote}\small
User [product]-shopping request:\\
Title: [title]\\{}
[body]\\
Below is the candidate list. Pick the SINGLE best recommendation.\\
Reply with EXACTLY one line:\\
CHOICE: <number>\\
Candidates:\\{}
[numbered candidate list]\\
Output only one line.
\end{quote}
Appendix~\ref{app:enum} specifies sampling, output limits, pool construction,
and choice parsing.

\paragraph{Query-lift prompts.}
Both conditions request \texttt{CHOICE: <number>} on the first line
and allow an explanation. The query-blind condition asks ``Which of
these is the best [product] in general?''; the conditioned one asks
``A user asks: [query]. Which of these is the best [product] for this
user?'' Both append the same numbered list. Thus query content and
the decision instruction change together.

\paragraph{ESCI prompt.}
The single user message requests \texttt{CHOICE: <number>} on the
first line, then asks ``A shopper searched for: [query]. Which product
best matches?'' followed by numbered truncated titles. Appendix~\ref{app:esci} specifies filtering, candidate sampling, title
truncation, and parsing. The historical records retain hit flags rather than
raw responses or selected product IDs.

\paragraph{Historical analyses.}
Earlier analyses used different reference sets, answer
extractors, sample aggregations, and candidate pools. Their numbers
are not pooled with this reanalysis. In particular, unpaired historical
root scores, the asymmetric body-class comparison, and title-only
concession aggregates do not support the main claims here.

\section{Reanalysis protocol and uncertainty estimation}
\label{app:reanalysis-protocol}

\paragraph{Objective-check scope and sources.}
The retrospective audit retains 149 shared car requests with an
explicit drivetrain or transmission requirement and all 3,129
saved answers (447 per model). These requests form the audited
subset of the 800-query corpus; the remaining 651 are outside its
scope. All 149 request annotations were reviewed against the
generation-visible text. One missing 4WD requirement was corrected
for all seven models. Automatic transmission includes CVT and
dual-clutch transmission; AWD and 4WD share one category.
All 279 nonempty extracted-name groups were screened for source-name
identity, and the final shared rules were applied to every answer.

US EPA records, Canadian fuel-consumption data, and manufacturer
specifications supply the configuration evidence. Year ranges,
body types, engines, turbo status, and recognized variants restrict
each name match. Uncovered years, ambiguous families, and unverified
trim tokens remain unresolved. Shared US-source rules require a
US purchase market established from the query; dollars or miles alone
are insufficient. The request review identified 55 US, 13 Canadian,
six other, and 75 unspecified markets. Individually reviewed cases
can use separately retained regional evidence. Source rows, identity
maps, request annotations, and response hashes are archived with
the final run.

\paragraph{Review and outcome accounting.}
The initial assistant review covered 159 full query/response pairs:
all 89 surfaced candidate/correction cases and a fixed sample of 70
apparent passes (ten per model). Source-evidence review covers one
further unresolved case, 130 cases reclassified as supported,
and five apparent supports reclassified as unknown. Complete-answer
checks cover all these cases. The apparent-pass sample is an audit,
not a random estimate of correctness. All surfaced flags received
a disposition. A confirmed conflict requires a source-backed
violation of a requested field that remains in the recommendation;
self-retractions, disclosed trade-offs, and ambiguous alternatives
are kept separate. The 356 supported configurations comprise 33
individually reviewed supports and 323 conservative source-rule
matches (Table~\ref{tab:objective-audit}). This support concerns the
primary recommendation's audited fields. It does not certify the
remaining statements, price, availability, or overall suitability.
The other/unknown category includes unresolved identities and
markets, unspecified configurations, corrected or rejected flags,
and disclosed trade-offs; none is counted as supported.

\begin{table*}[t]
\centering
\small
\begin{tabular}{@{}lrrrr@{}}
\toprule
Model & Reviewed support & Rule support & Conflict minimum (\%; 95\% CI) & Other/unknown \\
\midrule
GPT-5 & 9 & 71 & 0.00 [--] & 367 \\
Claude Opus & 5 & 39 & 0.00 [--] & 403 \\
Llama-70B & 8 & 54 & 1.34 [0.00, 3.36] & 379 \\
Qwen-80B & 3 & 54 & 0.67 [0.00, 2.01] & 387 \\
Nemotron-49B & 5 & 54 & 0.67 [0.00, 1.79] & 385 \\
Llama-8B & 2 & 34 & 1.12 [0.00, 2.91] & 406 \\
Mistral-14B & 1 & 17 & 3.80 [1.79, 6.27] & 412 \\
\bottomrule
\end{tabular}
\caption{\textbf{Evidence coverage and uncertainty for the objective audit.}
Each model has 447 answers from the same 149 requests. The two
support columns partition the supported counts in
Table~\ref{tab:objective}. Intervals use 2,000 whole-query bootstrap
resamples (seed 42), retaining all three generations together.
They describe the detected conflict statistic and exclude uncertainty
about conflicts hidden among unknown cases. Zero-event bootstrap
intervals are withheld (--); zero confirmed conflicts does not
establish a zero population rate.}
\label{tab:objective-audit}
\end{table*}

The assessable query/sample intersection across all seven models is
empty. We therefore retain every answer in the denominator and report
minimum confirmed conflicts, without ranking conditional rates on
different assessable subsets. This is retrospective assistant
adjudication, not independently annotated gold data. Source-certificate review
reclassifies 130 cases as supported and five as unresolved because their
year/trim scope exceeds the evidence, a net increase of 125 supported answers.
These five cases are evidence-scope abstentions, not detected drivetrain errors.
The cohort and confirmed-conflict counts are fixed throughout the audit.

\paragraph{Headphone connection audit.}
A request-only screen identified 44 of the 226 shared headphone requests with
explicit connection requirements in the generation-visible text: 25 wireless,
11 Bluetooth, and eight wired. All seven models contribute three saved answers
per request, giving 924 answers. One assistant reviewed the complete responses
for product identity, final recommendations, alternatives, and accessory bundles.
Model-blinded verification of manufacturer evidence covered 49 distinct wired
capability gaps and 33 exact-product nonwired gaps, establishing 52
product/capability facts; 42 ambiguous nonwired gaps remained outside that pass.
The evidence map and full-response review of 118 potentially affected answers
were frozen before scoring all 924 answers.

\begin{table*}[t]
\centering
\small
\begin{tabular}{@{}lrrrr@{}}
\toprule
Model & Supported & Conflicts & Unknown & Supported \% [95\% CI] \\
\midrule
GPT-5 & 125 & 0 & 7 & 94.7 [88.6, 99.2] \\
Claude Opus & 128 & 0 & 4 & 97.0 [93.9, 99.2] \\
Llama-70B & 110 & 0 & 22 & 83.3 [72.0, 93.2] \\
Qwen-80B & 127 & 0 & 5 & 96.2 [91.7, 100.0] \\
Nemotron-49B & 99 & 0 & 33 & 75.0 [65.2, 84.1] \\
Llama-8B & 106 & 0 & 26 & 80.3 [71.2, 88.6] \\
Mistral-14B & 100 & 0 & 32 & 75.8 [67.4, 83.3] \\
\bottomrule
\end{tabular}
\caption{\textbf{Headphone connection-capability audit.} Counts retain all
132 answers per model (44 requests, three responses). Supported percentages
reflect evidence coverage. Intervals use 10,000 request-cluster resamples
(seed 20261012), stratified by the two annotation batches of 20 and 24 requests.
Zero conflicts does not establish zero population risk or a model-quality ranking.}
\label{tab:headphone-objective}
\end{table*}

Manufacturer evidence supports the requested connection in 795 answers;
129 remain unknown, with no confirmed conflicts (Table~\ref{tab:headphone-objective}).
Ten requests are assessable across all models and generations, with all 210
answers supported. Fifty-four unknown answers use unverified HD4.50BT or
HD4.50BT ANC abbreviations. Although the search considered positive and explicit
negative capability statements, no exact-product negative was established;
missing evidence remains unknown. Supported counts measure evidence coverage,
so an increase in support under a broader source map is not a change in model
performance.

Annotation uses two batches of 20 and 24 requests; the 20-request batch was
selected by a fixed hash order before answer review. The bootstrap retains
these batches as strata. Annotations come from one assistant, not independent
human validation; intervals exclude annotation and source uncertainty.
The check covers only the top actionable recommendation's connection capability.
Price, sound quality, comfort, other requested features, and historical
availability remain unassessed. The attribute differs from the car audit and
cannot establish comparative domain difficulty.

\paragraph{Reanalysis protocol.}
We recomputed the Reddit agreement and matched enumeration results
from saved generations with one declared product matcher: the
token-boundary matcher with the alias and variant-suffix rules specified in Appendix~\ref{app:agreement}. The answer extractor scans a recommendation
section preferentially, then the response, using a catalog vocabulary
purged of fragments and designated family wildcards. It can still
mistake a mentioned alternative for the final pick.
For reference sensitivity we score against the first raw reference
entry and against all raw entries, keeping the same outputs and
denominator. All three saved samples contribute equally within each
query. Unextractable outputs remain in the denominator and score zero.
The protocol was specified retrospectively.

\paragraph{Comparisons and uncertainty.}
Enumeration coverage is measured at nested depths on the same queries;
its final choices and the free-form baseline use the same three-sample
mean. Enumeration consumes additional generation calls, so this is
not a cost-matched efficiency comparison. Query lift is paired within
query, with the query-blind prompt as control. ESCI uses uniform
four-way choice as a reference level.
We use 2,000 deterministic query-bootstrap resamples for the agreement
and enumeration tables and 10,000 for the extraction and baseline
sensitivity checks, with Wilson intervals for ESCI's binary rates.
Intervals describe query variation conditional on the saved outputs,
not variance across generation seeds.

 \section{Full-request source-blind comparison}
\label{app:preference}

\paragraph{Sample and fixed alternative.}
The prespecified protocol
samples 150 car requests from 1,247 eligible source requests using
seed 20260927. Eligibility requires a nonempty full request and an
unambiguous catalog-recognizable community mention. The first such
entry in the stored reference order supplies one fixed alternative
per request, selected before model generation. The original answer can itself match another
community mention; eligibility for the comparison concerns only the
fixed alternative.

\paragraph{Generation and comparison.}
All six models receive the same title, full body, and historical-date
instruction. Initial generations use temperature 0.7, seeds 42--44,
and a 2,048-token output limit. Each must return JSON with one
recommendation and an explanation, or abstain. Initial abstentions,
invalid responses, and recommendations matching the fixed alternative
are ineligible; Table~\ref{tab:preference-attrition} accounts for all
450 initial generations per model. Each eligible trial has two fresh,
source-blind comparisons, at temperature zero and a 1,024-token output
limit. They present the same two name strings in opposite orders,
without the initial explanation or conversation history. The allowed
choices are A, B, EQUAL, NEITHER, and INSUFFICIENT. Scoring maps the
selected letter back to its supplied name; no judge interprets the
explanation to determine the winner. Invalid or missing choices cannot
count as a reference win and remain in the denominator. All scheduled
calls completed without terminal API failures; unsuccessful parsing
was retained rather than retried.

\begin{table*}[t]
\centering
\small
\begin{tabular}{@{}lrrrrrrr@{}}
\toprule
Model & Invalid & Abstain & Match & Excl. & Eligible & Bad comp. & Ref. / presentations \\
\midrule
GPT-5.4 & 8 & 1 & 39 & 0 & 402 & 0 & 77/804 (9.6\%) \\
Claude Opus 4.8 & 14 & 3 & 36 & 5 & 392 & 1 & 53/784 (6.8\%) \\
Llama-70B & 13 & 0 & 25 & 7 & 405 & 6 & 70/810 (8.6\%) \\
Qwen-80B & 5 & 3 & 36 & 6 & 400 & 1 & 58/800 (7.2\%) \\
Nemotron-49B & 91 & 16 & 21 & 5 & 317 & 2 & 70/634 (11.0\%) \\
Llama-8B & 37 & 11 & 25 & 10 & 367 & 26 & 120/734 (16.3\%) \\
\bottomrule
\end{tabular}
\caption{\textbf{Source-blind comparison eligibility and per-presentation results.}
Each model starts with 450 initial generations (150 requests, three seeds).
The first five numeric columns partition these generations: invalid
initial responses, abstentions, parser-detected reference matches,
post-hoc identity exclusions, and retained eligible trials.
Invalid comparisons (``Bad comp.'') remain in the denominator of
two presentations per retained trial. The final column counts
reference choices in individual presentations, including order-sensitive
choices, with percentages in parentheses.}
\label{tab:preference-attrition}
\end{table*}

\paragraph{Complete prompts and response formats.}
The initial generation receives a single user message below;
bracketed fields are replaced with the stored request date, title,
and full body. No system message is added.
\begin{quote}
You are a car recommendation advisor. Read the user's request
and give one practical recommendation that fits the stated needs.

Return only a JSON object with ``recommendation'' (one specific
car make and model, or null if there is insufficient information) and
``reason'' (a brief explanation, at most 120 words). Do not list
alternatives in the recommendation field.

Treat the request as made on [request date]. Use only vehicles
available by that date.\\
User's request:\\
Title: [title]\\
{}[full body]
\end{quote}
Thus, the initial response has the form
\texttt{\{"recommendation": "[car name]", "reason": "[explanation]"\}},
with JSON \texttt{null} replacing the car-name string for abstention.

Each source-blind comparison also receives just one user
message, without the initial response or conversation history:
\begin{quote}
Treat the request as made on [request date]. Use only vehicles
available by that date.\\
User's request:\\
Title: [title]\\
{}[full body]

Which is the better recommendation for this request? The
options are unlabeled as to source. If they are equally suitable,
neither is suitable, or the information is insufficient, report that
rather than force a preference.\\
A. [first car name]\\
B. [second car name]

Return only JSON with ``choice'': ``A'', ``B'', ``EQUAL'',
``NEITHER'', or ``INSUFFICIENT'', and ``reason'': a brief explanation
(at most 120 words).
\end{quote}
The comparison response has the form
\texttt{\{"choice": "[allowed choice]", "reason": "[explanation]"\}}.
One ordering places the initial recommendation first and the fixed
alternative second; the other reverses them. Source labels are absent
to avoid explicitly cueing deference to the community or defense of
the model's earlier answer. The execution order is deterministically
shuffled within each trial; all prompts and raw responses are saved.

\paragraph{Identity sensitivity check.}
For GPT-5.4 and Claude, inspection of all eligible
initial-name/reference pairs found three
Claude trials comparing ``Volkswagen Golf GTI'' with ``Volkswagen GTI'';
both blind responses identify the names as equivalent. Two other
Claude recommendation fields contain explicit alternatives: ``Mazda
MX-5 Miata (or Mazda3 for practicality)'' and ``2019 Mazda3 Sedan
(or Hatchback).'' The original parser missed the parenthesized
disjunctions. We conservatively exclude all five trials, including
the within-model body-style alternative, leaving 392 Claude trials.
None is a reference win under either the both-orders or per-presentation metric; the
numerators do not change. No GPT-5.4 trial is excluded. Both retained
cohorts cover 137 of the 150 sampled requests, not necessarily the
same 137 requests.

These exclusions form a post-hoc sensitivity analysis.
Before exclusions, the primary results are 26/402 (6.5\%) and
16/397 (4.0\%); per-presentation results are 77/804 (9.6\%) and
53/794 (6.7\%). The checked figures in the main text use 402 and
392 as trial denominators for these two models. The check verifies saved prompts, parsing,
option-order mapping, and outcome counts, and retains all 42 positive
trial records with their explanations. It does not certify those
explanations as correct product facts or independently validate
recommendation quality.

The same conservative identity check excludes seven Llama-70B,
six Qwen, five Nemotron, and
10 Llama-8B trials: 22 overlapping name/family/trim identities,
two explicit alternative configurations, three ``Subaru dealer''
responses, and one conflicting ``Subaru CX-5'' make/model string.
Overlaps include Miata versus MX-5 Miata, BMW 3-Series versus 328i,
and abbreviated Jetta, Bolt, or 4Runner names. None is a both-orders
reference win. The frozen primary counts are 26/412, 18/406,
13/322, and 7/377; after exclusions they are 26/405, 18/400,
13/317, and 7/367. Their 64 positive trials span 10, 13,
seven, and six requests, respectively. All 106 positive trials
across the six models are retained with their explanations in the
derived audit.

Eligibility also differs across models. Nemotron has 91/450
invalid initial responses (20.2\%) and 16 abstentions; Llama-8B
has 37 invalid responses (8.2\%) and 11 abstentions. The other
four models have 5--14 invalid initial responses each (1.1--3.1\%).
These losses include JSON/schema violations and multiple-name fields,
limiting the comparison's coverage.

\paragraph{Uncertainty and archived controls.}
Intervals use 5,000 bootstrap resamples of eligible request IDs with
seed 20260927, retaining each request's initial generations and both
orders together. They describe variation across these requests
conditional on the saved outputs, not independent evidence from each
presentation. The GPT/Claude car collection also includes two reconsideration
conditions with the original answer visible and an otherwise matched
no-alternative reconsideration control. These saved follow-ups are
not used in the source-blind headline. In particular, the control's
raw change count can include year- or trim-only changes and must not
be read as a rate of changes in the recommended car model.

\paragraph{Collection and audit.}
The GPT/Claude car collection requests
\path{openai/openai/gpt-5.4} and \path{azure/anthropic/claude-opus-4-8}.
Returned model identifiers are recorded for every response. Primary and
identity-screened results use the same saved responses; the latter apply only
the five exclusions listed above. The earlier title-only concession analysis
is not pooled with this run: cohort, generation, date instruction, and response
format differ.

The open-model car collection was recorded on October 10, 2026, using the same
frozen cases, prompts, seeds, and token limits. It evaluates Llama-3.3-70B-Instruct,
Qwen3-Next-80B-A3B-Instruct, Llama-3.3-Nemotron-Super-49B-v1,
and Llama-3.1-8B-Instruct through a hosted inference endpoint.
The requested Ministral-14B-Instruct-2512 version was unavailable;
no replacement version was substituted. The failed public-endpoint
probes are infrastructure records and are not scored. This collection
contains 1,800 initial responses and 3,034 blind comparisons, with
no conversational revision controls. Collection dates and serving
implementations differ from the GPT/Claude collection. We checked every
prompt and scored outcome against the prespecified protocol and report primary
and identity-screened results separately.

\subsection{Temperature-zero initial recommendations}
\label{app:temperature-zero}

\paragraph{Protocol.}
The three-sample experiment uses initial recommendations at temperature 0.7;
a sampled answer need not be the model's preferred recommendation. We therefore
generated one initial answer at temperature zero for each of the same 150 car
requests, using GPT-5.4 and Opus 4.8 with nominal seed 42. We retained the full
requests, dates, fixed alternatives, prompts, token budgets, parser, and
requested model routes. Valid disagreements received both source-blind orders
at temperature zero. The check used 414 GPT and 412 Opus calls, all successful
on their first transport attempt. Malformed outputs were retained.
Temperature zero removes deliberate sampling randomness but does not guarantee
deterministic API serving.

\paragraph{Eligibility and review.}
Of 150 initial GPT answers, three fail the inherited name schema and 15 match
the reference, leaving 132 disagreements. For Opus, eight initial answers fail
parsing, one abstains, and 10 match, leaving 131. The identity sensitivity
excludes two further Opus pairs: Golf GTI versus GTI (duplicate labels), and
GTI versus an unspecified Golf (family/variant overlap). Neither is a
both-order reference win. The retained denominator is 129, including one
invalid comparison pair; valid-pair sensitivity gives 4/128 (3.1\%).
GPT has no invalid comparison pair. All 300 initial-name pairs were reviewed,
together with the first 10 requests per model, all 11 positive cases, and
every malformed output. Parsing and identity screening were not treated as
objective-quality judgments.

\begin{table*}[t]
\centering
\small
\begin{tabular}{@{}lrrrrr@{}}
\toprule
Model & Initial temp. & Initial answers & Eligible & Both orders & Rate \% [95\% CI] \\
\midrule
GPT-5.4 & 0.7 & 450 & 402 & 26 & 6.5 [3.0, 10.3] \\
GPT-5.4 & 0 & 150 & 132 & 7 & 5.3 [2.3, 9.1] \\
Opus 4.8 & 0.7 & 450 & 392 & 16 & 4.1 [1.5, 7.3] \\
Opus 4.8 & 0 & 150 & 129 & 4 & 3.1 [0.8, 6.2] \\
\bottomrule
\end{tabular}
\caption{\textbf{Temperature-zero robustness of both-order community preference.}
Both orders counts requests/trials choosing the fixed community alternative
in both presentations, after identity screening. The 0.7 rows use three initial
generations per request; the zero rows use one. Rates divide by eligible
disagreements; intervals use 5,000 whole-query bootstrap resamples
(seed 20260927). Initial products, eligible cohorts, and collection dates
differ across temperatures.}
\label{tab:temperature-zero}
\end{table*}

\paragraph{Results and interpretation.}
The alternative wins in both orders for 7/132 GPT requests (5.3\%; 95\%
interval 2.3--9.1\%) and 4/129 Opus requests (3.1\%; 0.8--6.2\%;
Table~\ref{tab:temperature-zero}). These point estimates are lower than the
temperature-0.7 rates of 6.5\% and 4.1\%, but the behavior remains observable without
deliberately sampling the initial answer. This supports a limited claim about
disagreement between generation and comparison, not the objective superiority
of the alternative. The different initial products, eligibility sets, and
collection dates preclude attributing the rate difference solely to temperature.
Some positive-case explanations introduce unstated priorities or questionable
product claims; choosing the reference does not verify the reasoning.
Per presentation, the reference wins 24/264 GPT and 22/258 Opus comparisons;
order sensitivity remains possible.

Primary, identity-screened, and valid-pair
analyses use the same saved responses without further generation. Review was
by one assistant, not independent human annotation.

\subsection{Restricted laptop replication}
\label{app:laptop-preference}

\paragraph{Cohort and protocol.}
We extended the source-blind comparison to GPT-5.4 and Claude Opus 4.8
on the same 150 laptop requests, sampled with seed 20261010 from
201 eligible requests among 961 stored posts. Eligibility required
a full dated request and an exact catalog manufacturer/model-series
name also present in a stored comment. The first eligible reference
in stored order was fixed before generation. Broad families and
underspecified names were excluded. Of the 150 fixed references,
80 name MSI, 57 Dell, nine Lenovo, and four ASUS products, on requests
posted from January 1 to March 8, 2019.

The prompts retain the full request and historical-date instruction.
Three initial generations use seeds 42--44 and temperature 0.7;
each eligible disagreement receives both source-blind orders at
temperature zero. Initial agreement uses manufacturer/model-series
identity. The outcome is preference for the reference in both orders
per initial generation, without voting across the three generations.

\paragraph{Parser and identity audit.}
For GPT, review of 18 initially excluded names recovered 16
single-product names containing configuration details: three matched
the reference, and 13 received their 26 missing comparisons.
The 450 initials then comprise 46 same-series matches, two exclusions,
and 402 disagreements. A separate identity sensitivity excludes two
probable Dell G5 aliases, leaving 400 trials across 142 requests.

For Opus, review recovered 17 single-product initials and their
34 missing comparisons, leaving 433 valid initials: 23 same-series
matches and 410 disagreements. The remaining 17 comprise 14
multiple/invalid names, two format errors, and one terminal transport
failure. Seven further uncertain identities are excluded only in the
sensitivity analysis, leaving 403 trials across 143 requests.
Six malformed comparison responses remain in the parser-corrected
denominator; five remain after identity exclusions. GPT has no
malformed comparisons. Neither initial answers nor completed
comparisons were regenerated during recovery.

\begin{table*}[t]
\centering
\small
\begin{tabular}{@{}llrr@{}}
\toprule
Model & Analysis & Preferences / trials & Rate (\%; 95\% CI) \\
\midrule
GPT-5.4 & Frozen parser & 43 / 389 & 11.05 [6.57, 15.97] \\
GPT-5.4 & Parser-corrected & 44 / 402 & 10.95 [6.52, 15.83] \\
GPT-5.4 & Identity sensitivity & 44 / 400 & 11.00 [6.55, 15.88] \\
Claude Opus 4.8 & Frozen parser & 42 / 393 & 10.69 [6.44, 15.25] \\
Claude Opus 4.8 & Parser-corrected & 43 / 410 & 10.49 [6.37, 14.88] \\
Claude Opus 4.8 & Identity sensitivity & 41 / 403 & 10.17 [6.09, 14.64] \\
\bottomrule
\end{tabular}
\caption{\textbf{Preference for community-mentioned laptop alternatives.}
Both-order preferences per eligible initial generation, on the same
restricted 150-request cohort. Rows within each model are successive
analyses of one run. Intervals use 5,000 whole-query bootstrap
resamples (seed 20260927); malformed comparisons remain in the
eligible denominator.}
\label{tab:laptop-preference}
\end{table*}

\paragraph{Results and interpretation.}
After identity exclusions, GPT prefers the reference in 44/400 trials
(11.0\%) and Opus in 41/403 (10.2\%;
Table~\ref{tab:laptop-preference}). On 368 commonly eligible
query/seed trials across 136 requests, Opus has 38 both-order
preferences and GPT 37: a paired difference of $+0.27$ percentage
points (95\% CI $[-5.98,+6.16]$), without established model separation.
The models generated different original alternatives, so even this
paired comparison concerns the joint generation/comparison procedure.
Opus's identity-filtered valid-pair rate is 41/398; allowing every
unresolved comparison pair that could favor the reference gives
42/403.

Assistant review covered the fixed 20-request samples, valid
initial/reference pairs, positive trials, and their source contexts.
References include criticized products, retracted suggestions,
and affiliate-style links. Some positive comparisons impute
configurations absent from the series-only reference. No post-hoc
endorsement filter changes the frozen cohort. These are preferences
for source-mentioned alternatives of uneven specificity, not
independent evidence that the original advice was wrong.
Differences from cars cannot be attributed to domain alone.

\paragraph{Collection.}
Including the additional comparisons after parser recovery, the GPT replication
used 1,254 successful calls to \path{openai/openai/gpt-5.4}.
The Opus replication used 1,269 successful calls to
\path{azure/anthropic/claude-opus-4-8}, plus the terminal initial failure.
Product facts in explanations and historical availability remain unverified.

\section{Agreement reanalysis and extraction coverage}
\label{app:agreement}

\paragraph{Support for the top reference.}
In the larger car reference corpus, 1,202 of 1,465 requests
(82.0\%) have a top reference name extracted from just one stored
top-level comment. Thus, the highest stored reference weight often
comes from a single comment. Counting distinct endorsing users would
require author IDs and endorsement judgments, which are unavailable.

\paragraph{Agreement scoring.}
The comparison uses all 800/800/226 saved query IDs in each Reddit
domain for every model, with three outputs per query. We verified
that query sets agree across models. The extraction vocabulary excludes entries whose model names contain only
generic/function words, designated plural family entries, and short alphabetic
line prefixes shared by multiple numbered models. Reference strings themselves
remain raw for the reference-granularity comparison. For matching, we split
letter--digit boundaries, lowercase names, replace non-alphanumeric characters
with spaces, and remove a final s from tokens longer than three characters.
A match requires contiguous whole-token containment in either direction.
Containment ending in a numeric token is rejected if the longer name continues
with a single-letter variant suffix, distinguishing T14 from T14s. If neither
name is a contiguous token subsequence of the other, an alias match can apply:
both names must share the first token, and the shorter name must have at least
two further tokens, at least one containing a digit, all present in the longer
name after its first token. These rules can award family-level credit and can
conflate abbreviated sibling products; the reported identity sensitivities
assess some of these cases.

For each query, we average the three binary hit outcomes before
averaging across queries. No-pick outputs remain zero in both scores.
The paired set-minus-top comparison changes no generation, extractor,
query inclusion rule, or sample aggregation. Table~\ref{tab:agreement-ci}
also reports extraction success, which varies materially by domain
and model. A missing extraction can occur even when the response
contains a useful recommendation.

\begin{table*}[t]
\centering
\small
\setlength{\tabcolsep}{3pt}
\begin{tabular}{@{}llrrrr@{}}
\toprule
Domain & Model & Top [95\% CI] & Set [95\% CI] & Difference [95\% CI] & Parsed (\%) \\
\midrule
Cars & GPT-5 & 6.8 [5.2, 8.5] & 24.8 [22.0, 27.6] & 18.1 [15.6, 20.6] & 95.3 \\
Cars & Claude Opus & 6.5 [5.0, 8.1] & 24.0 [21.4, 26.8] & 17.6 [15.1, 20.2] & 93.5 \\
Cars & Llama-70B & 6.0 [4.6, 7.5] & 24.5 [21.7, 27.0] & 18.5 [16.2, 20.9] & 96.7 \\
Cars & Qwen-80B & 5.2 [3.8, 6.8] & 23.2 [20.6, 26.0] & 18.0 [15.6, 20.4] & 97.5 \\
Cars & Nemotron-49B & 6.0 [4.6, 7.4] & 22.0 [19.5, 24.6] & 16.1 [13.9, 18.4] & 97.5 \\
Cars & Llama-8B & 5.6 [4.2, 7.0] & 21.3 [18.8, 24.0] & 15.7 [13.5, 18.1] & 95.8 \\
Cars & Mistral-14B & 5.5 [4.2, 6.8] & 20.8 [18.5, 23.3] & 15.4 [13.2, 17.5] & 98.1 \\
\addlinespace
Laptops & GPT-5 & 10.3 [8.5, 12.3] & 26.0 [23.4, 28.6] & 15.7 [13.6, 18.0] & 81.1 \\
Laptops & Claude Opus & 9.6 [7.8, 11.4] & 24.8 [22.2, 27.4] & 15.2 [13.0, 17.5] & 80.6 \\
Laptops & Llama-70B & 10.0 [8.2, 12.0] & 27.4 [24.6, 30.4] & 17.3 [15.0, 19.8] & 73.1 \\
Laptops & Qwen-80B & 5.0 [3.7, 6.4] & 19.0 [16.6, 21.5] & 14.0 [11.9, 16.1] & 72.6 \\
Laptops & Nemotron-49B & 8.6 [6.9, 10.2] & 23.3 [21.0, 25.8] & 14.8 [12.6, 17.0] & 73.5 \\
Laptops & Llama-8B & 9.0 [7.3, 10.7] & 24.0 [21.5, 26.6] & 15.0 [12.9, 17.2] & 70.6 \\
Laptops & Mistral-14B & 5.1 [3.9, 6.5] & 18.4 [16.1, 20.8] & 13.2 [11.3, 15.2] & 81.7 \\
\addlinespace
Headphones & GPT-5 & 6.6 [3.8, 9.9] & 22.6 [17.8, 27.4] & 15.9 [11.7, 20.6] & 87.3 \\
Headphones & Claude Opus & 7.2 [4.3, 10.3] & 24.6 [19.5, 29.8] & 17.4 [12.7, 22.1] & 91.7 \\
Headphones & Llama-70B & 8.3 [5.2, 11.5] & 21.5 [16.8, 26.4] & 13.3 [9.4, 17.1] & 86.1 \\
Headphones & Qwen-80B & 5.5 [2.8, 8.6] & 17.1 [12.7, 21.5] & 11.7 [8.3, 15.6] & 90.9 \\
Headphones & Nemotron-49B & 4.9 [2.8, 7.2] & 17.7 [13.7, 21.8] & 12.8 [9.3, 16.8] & 89.7 \\
Headphones & Llama-8B & 5.2 [2.9, 7.7] & 16.1 [12.5, 20.1] & 10.9 [7.5, 14.3] & 81.6 \\
Headphones & Mistral-14B & 3.7 [1.9, 5.8] & 14.7 [11.2, 18.4] & 11.1 [8.3, 14.0] & 96.2 \\
\bottomrule
\end{tabular}
\caption{\textbf{Agreement uncertainty and extraction coverage.}
Percentages with 95\% query-bootstrap intervals; differences are
paired set-minus-top percentage points. Parsed is the fraction of
saved outputs yielding a catalog-extracted pick. Every sample remains
in the agreement denominator.}
\label{tab:agreement-ci}
\end{table*}

\paragraph{Scope of the rescore.}
Reference endorsement, final-pick extraction, and historical availability
remain separate validation questions. The older catalog-validated
top-three metric used a different reference target. The former Stack
Exchange baseline counted mentions in a response window;
Table~\ref{tab:agreement-ci} uses one extracted pick on the Reddit
domains. The software analysis below examines extraction scope;
Appendix~\ref{app:lift} separately reports supplied-choice results.

\paragraph{Software extraction-scope sensitivity.}
A separate recovery uses 400 shared Software Recommendations requests
and three saved responses from each of seven models (8,400 answers).
Its 651 stored reference entries were traced to source answers and
the two consensus extractors. We compare literal reference-name
occurrence in the extracted opening recommendation label with
occurrence anywhere in the complete answer. All 1,200 answers per
model remain in the denominator, including unresolved opening labels.

\begin{table*}[t]
\centering
\small
\setlength{\tabcolsep}{4pt}
\begin{tabular}{@{}lrrrr@{}}
\toprule
Model & Opening (\%) & Full answer (\%) & Difference (pp) & Unparsed \\
\midrule
GPT-5 & 24.3 [20.3, 28.4] & 45.9 [41.3, 50.5] & 21.7 [18.4, 25.3] & 5 \\
Claude Opus & 20.8 [17.1, 24.8] & 44.7 [40.1, 49.4] & 23.8 [20.2, 27.7] & 48 \\
Llama-70B & 23.2 [19.3, 27.2] & 40.4 [35.9, 44.8] & 17.3 [14.2, 20.6] & 2 \\
Qwen-80B & 20.9 [17.3, 24.8] & 41.9 [37.3, 46.5] & 21.0 [17.7, 24.5] & 36 \\
Nemotron-49B & 17.5 [14.3, 20.8] & 37.4 [33.2, 41.8] & 19.9 [16.9, 23.1] & 60 \\
Llama-8B & 14.6 [11.8, 17.6] & 33.0 [28.9, 37.3] & 18.4 [15.5, 21.5] & 126 \\
Mistral-14B & 14.2 [11.3, 17.3] & 39.4 [35.0, 43.9] & 25.3 [21.9, 28.8] & 59 \\
\bottomrule
\end{tabular}
\caption{\textbf{Software reference-name occurrence by extraction scope.}
Percentages and paired percentage-point differences with 95\%
intervals from 4,000 query-cluster resamples (seed 20261010).
Unparsed counts unresolved opening labels among 1,200 answers per
model. A name occurrence can be an alternative, component, or
rejection; these are not semantic recommendation-accuracy scores.}
\label{tab:software-extraction}
\end{table*}

Full-answer occurrence exceeds opening-label occurrence by
17.3--25.3 points (Table~\ref{tab:software-extraction}). On the
912 query/sample positions parsed for every model, gaps remain
16.8--25.2 points. Excluding 18 entries with generic names leaves
17.1--25.0-point gaps on the full denominators. These sensitivities
support an extraction-scope effect, while composite recommendations,
aliases, and retractions still require semantic judgment.
The parser was developed on disclosed subsets and received assistant
spot checks; this is not an independent human extraction benchmark.
The source-order cohort, edited source text, and incomplete serving
provenance also limit interpretation.
\section{Failure taxonomy and assessment requirements}
\label{app:taxonomy}

Table~\ref{tab:taxonomy} lists all 19 conceptual modes and the evidence
needed to establish each diagnosis. For example, diagnosing an
unavailable candidate requires checking product availability beyond
the local catalog.

\begin{table*}[t]
\centering
\small
\setlength{\tabcolsep}{4pt}
\begin{tabular}{@{}p{0.8cm}p{4.1cm}p{9.2cm}@{}}
\toprule
Code & Intended failure mode & Evidence required for assessment \\
\midrule
1a & A relevant fact is omitted & Request spans, recovered facts, and a relevance criterion. \\
1b & A fact is invented & The asserted fact and evidence that the request does not support it. \\
1c & Valence or attribution is changed & Who expressed the preference, its target, and its polarity. \\
2a & An inference is unsupported & The inferred belief and the facts claimed to license it. \\
2b & A prior is presented as user-derived & Provenance of the belief and an explicit distinction between evidence and assumption. \\
2c & A prior contradicts the request & A contradiction between an added belief and a supported user fact. \\
3a & A constraint is omitted & An explicit or justified requirement and its representation in $C$. \\
3b & A constraint has the wrong value & The source value, units, and represented value. \\
3c & Hard and soft criteria are confused & Language establishing requirement versus preference; ambiguous cases may remain unknown. \\
3d & Infeasibility is not acknowledged & Evidence that the specified constraints cannot jointly be met in the relevant option space. \\
3e & An allowed class is prematurely excluded & A request allowing alternatives, and evidence that a class was removed without justification. \\
4a & Candidate coverage is inadequate & An explicit pool and a declared target or breadth criterion; a small pool alone is a proxy. \\
4b & A candidate is nonexistent or unavailable & Reliable product identity and availability at the evaluation place and date. \\
4c & A candidate belongs to the wrong tier & The requested tier and relevant product attributes, with units and temporal context. \\
5a & A satisfying candidate is discarded & The original and retained pools plus assessable constraints showing a false negative. \\
5b & A violating candidate is retained & The retained option and an assessable violated hard constraint. \\
6a & Ranking conflicts with stated priorities & The stated preferences, relevant attributes, and a defensible comparison rule. \\
6b & The final pick contradicts the explanation & An identifiable pick and a contradiction in the response's own explanation. \\
6c & Choice varies under equivalent conditions & Repeated, matched inputs and valid parsed selections; variation alone need not be an error. \\
\bottomrule
\end{tabular}
\caption{\textbf{The 19-mode diagnostic taxonomy.} The last column
specifies the evidence needed for each diagnosis.}
\label{tab:taxonomy}
\end{table*}

\paragraph{Illustrative use.}
Consider a constructed request for a six-seat family vehicle, with a
fixed budget, allowing either an SUV or a minivan. If the extracted
constraints require an SUV, mode 3e directs attention to the lost
disjunction. If the constraints preserve both classes but the pool
contains only SUVs, pool construction merits inspection; an SUV-only
list is not itself an error if its options satisfy the request. If a
six-seat candidate is removed despite satisfying the stated criteria,
the error lies in filtering. Each diagnosis requires different evidence.

\section{Prototype implementation versus intended checks}
\label{app:implementation}

Table~\ref{tab:impl} documents the selected prototype checks. Fifteen
checks are non-generative; four call a generative judge. Selection on
small constructed fixtures does not establish precision or recall on
natural responses. Several checks expect explicit \texttt{Step}
headings and are not general free-form extractors.

\begin{table*}[t]
\centering
\small
\setlength{\tabcolsep}{4pt}
\begin{tabular}{@{}p{0.8cm}p{4.7cm}p{8.6cm}@{}}
\toprule
Code & Selected implementation & Scope or qualification \\
\midrule
1a & Regular-expression coverage & Limited to canonical query patterns. \\
1b & Token overlap & Measures lexical overlap; semantic support requires separate assessment. \\
1c & Polarity and target lexicons & Limited sentiment and attribution coverage. \\
2a & Structural citation check & Inference validity requires checking the cited evidence. \\
2b & Strict provenance-tag pattern & Measures explicit labeling of assumptions. \\
2c & Generative judge & Requires judge calibration and records of its outputs. \\
3a & Expanded constraint patterns & Limited to recognized constraint forms. \\
3b & Value-pattern comparison & Sensitive to parsing, units, and scope. \\
3c & Modal-language patterns & A proxy for hard/soft distinctions. \\
3d & Infeasibility-flag phrases & Joint catalog feasibility requires separate assessment. \\
3e & Multi-class patterns & Limited to recognized alternative-class expressions. \\
4a & Candidate-count threshold & Selected recognizer is vehicle-specific; cross-domain firing is not a valid breadth estimate. \\
4b & Generative judge & Catalog absence and real-world nonexistence must be distinguished. \\
4c & Catalog price-tier comparison & Depends on accurate identity, price metadata, and the requested market. \\
5a & Broader constraint/mention extraction & Requires both pools and assessable constraints. \\
5b & Constraint patterns and mention scan & Same prerequisites; unknown attributes cannot count as violations. \\
6a & Generative judge & Requires a specified interpretation of preference trade-offs. \\
6b & Generative judge & Targets explanation--choice consistency. \\
6c & Hedge-word count & Not a measurement of cross-rollout agreement. \\
\bottomrule
\end{tabular}
\caption{\textbf{Selected prototype implementations.} The final column
states the scope and limitations of each heuristic or judge-based check.}
\label{tab:impl}
\end{table*}

\paragraph{Worked example: numerical check for failure 3b.}
In the constructed example in Section~\ref{sec:taxonomy}, the request
requires at least six passengers under \$20,000. The response records
both requirements correctly in $F$, but its hard-constraint section
$C$ specifies at least five passengers and price at most \$20,000.

The implemented 3b verifier extracts numbers by dimension from the
request and the response's labeled hard-constraint section. It checks
whether a represented value is within 10\% of the requested value.
Here the budget matches, but five differs from six by 16.7\%, so the
verifier flags the response and reports a score of 0.5: one of two
dimensions matches. Replacing five with six makes the score 1.0 and
clears the flag. We reproduced both outcomes with the implementation.

The numerical tolerance is a prototype heuristic. Compliance must be
assessed against the user's exact requirement.
The check also depends on recognizing the wording and section headers.

\paragraph{Root-check experiment.}
The five non-generative root checks are rerun on the original
Llama-70B root experiment's first 100 request IDs, using the same IDs
for every model and the first saved generation. The historical
analysis used the first 100 saved responses per model; their request order differs, so those cohorts were not fully paired. We reproduce those
historical car scores separately, then align the IDs and restrict the
query body used by each check to the generator's 1,800-character
limit. Table~\ref{tab:roots} reports the resulting matched profiles.
No generation is dropped for a verifier failure. The counts describe
this fixed sample and one generation per request.

The 1a check fires when any recognized query token is absent
from the parsed fact list. The 1b check fires when any fact item has
less than 50\% content-token overlap or contains a number absent
from the query. The 1c check searches for limited target/polarity
patterns. The 2b check requires a recognizable Step 2B section and
explicit prior tags on its items. The 4a check flags fewer than five
recognized vehicle names, except when the query already names at
least two vehicles. It uses the parsed post-filter pool when available
and otherwise scans the response. Its input can therefore differ from
the initial candidate pool.

Missing parseable facts cause 1a and 1b to return a non-firing
score, and queries without recognized targets can also pass 1c
without a substantive assessment. Conversely, harmless paraphrases
can trigger the lexical checks. A non-firing flag therefore does
not certify that the corresponding reasoning step is correct.
We do not restore the historical 100\% laptop/headphone 4a rates:
the selected recognizer is vehicle-specific, making those figures
out-of-domain extraction artifacts rather than candidate-coverage
measurements.

\paragraph{Exploratory checker--assistant comparison.}
We selected 20 of the shared 100 car requests by a fixed hash,
retaining the first response from all seven models (140 answers).
A request-only fact inventory and rubric were frozen before answer
annotation. One assistant read complete responses under opaque IDs;
its labels were frozen before model keys or checker outputs were
opened. Paraphrases counted as supported, and labeled assumptions
were distinguished from claims about the user's stated facts.

\begin{table*}[t]
\centering
\small
\setlength{\tabcolsep}{4pt}
\begin{tabular}{@{}lrrrrrr@{}}
\toprule
Root & Both flag & Checker only & Assistant only & Both clear & Unknown & Agreement (\%) \\
\midrule
1a & 63 & 17 & 25 & 33 & 2 & 96/138 (69.6) \\
1b & 24 & 77 & 2 & 20 & 17 & 44/123 (35.8) \\
\bottomrule
\end{tabular}
\caption{\textbf{Lexical checks versus exploratory assistant labels.}
Counts among 140 responses. Roots 1a and 1b concern omissions and
unsupported additions in the explicit user-facts stage.
Agreement excludes unknown assistant labels. These single-assistant
labels are not independently validated human gold.}
\label{tab:root-calibration}
\end{table*}

Agreement on definite cases is 69.6\% for 1a and 35.8\% for 1b
(Table~\ref{tab:root-calibration}). Disagreements include supported
paraphrases, equivalent number formats, and changed relationships
hidden by reused vocabulary. One readable fact list is missed by
the legacy parser, so agreement can also conceal a missing check.
Of 88 assistant-labeled stage omissions, 17 are fully recovered
elsewhere in the answer, 69 retain a whole-answer omission, and two
have uncertain recovery. All 280 saved checker outputs were
reproduced exactly. This small, subjective audit
motivates detector validation; it does not establish population
failure rates or a model ranking.

\paragraph{Fact-anchoring experiment.}
The fact-anchoring experiment
takes the first 400 eligible car requests and generates three
Llama-70B responses per request using seeds 42--44, temperature 0.7,
and a 1,100-token output limit, as in the structured baseline.
The prefilled fact list is formed from the title and body, capped
jointly at 2,500 characters, split on sentence boundaries and
newlines, retaining the first 12 segments longer than 12 characters.
The title and up to 4,000 body characters are also included in the
request, versus 1,800 body characters in the baseline. The model can
revise or omit the prefilled fact list in its response.

We match all 400 request IDs, verify the stored references
agree across conditions, and apply the same purged car catalog,
final-pick extractor, and name matcher to both sides. Each query's
score is the mean of its three binary sample scores; an unextractable
pick contributes zero. The all-entry agreement difference is
$-0.9$ points, with a paired query-bootstrap interval of
$[-2.8,+0.8]$. The post-hoc input-length sensitivity check excludes
the 39 requests with bodies longer than 1,800 characters, without
conditioning on outcomes. Table~\ref{tab:fact-anchor} reports the
remaining 361 requests. Intervals use 2,000 paired query resamples
with fixed, hash-derived seeds and retain the three generations
together. They quantify variation across requests conditional on
the saved outputs, excluding uncertainty from collection dates or
serving implementations.

\begin{table*}[t]
\centering
\small
\begin{tabular}{@{}lrrr@{}}
\toprule
Metric & Baseline & Anchored & Difference [95\% CI] \\
\midrule
Top-entry agreement & 5.1 & 4.6 & -0.5 [-1.5, +0.6] \\
All-entry agreement & 22.2 & 20.2 & -1.9 [-4.0, -0.2] \\
Pick extracted & 97.0 & 96.1 & -0.9 [-2.1, +0.2] \\
\bottomrule
\end{tabular}
\caption{\textbf{Fact anchoring on input-length-matched requests.}
Llama-70B, 361 paired car requests, three generations per condition.
Both prompts receive the same request content; the anchored prompt
additionally supplies the fact-list scaffold. Rates are percentages;
differences are anchored minus baseline percentage points with
95\% paired query-bootstrap intervals. Missing picks remain in
the agreement denominator.}
\label{tab:fact-anchor}
\end{table*}

Among the 1,083 anchored outputs in this subset, 82 lack
a parseable returned fact list, compared with none in the baseline.
The absence may reflect using the supplied list without restating it;
it cannot establish that the facts were ignored. Because the lexical
checks can return non-firing scores when this section is absent,
a reduction in flags would not demonstrate a successful fact repair.
Extraction rates across all 400 requests are 97.0\% and 96.1\%, and the
matched-subset rates are reported in Table~\ref{tab:fact-anchor}.

\paragraph{Missingness semantics.}
For an option-level constraint check, an unrecognized product or
missing attribute yields \emph{unknown}. For the claim that at least
one reference option satisfies a constraint, the result is true if
any is known to satisfy it, false only if all are assessable and
violating, and unknown otherwise. We
exclude the legacy body-class comparison from the empirical claims
because its treatment of unresolvable community options was asymmetric.

\section{Candidate-coverage details and controls}
\label{app:enum}

\paragraph{Recovered coverage across models.}
Table~\ref{tab:coverage-six} recovers six-model results from the saved
enumeration pools. We intersect the original 171/176/49 request IDs
with availability in all six models, yielding 160/164/49 shared IDs;
the exclusions arise from missing Llama-8B records.
Raw reference lists agree across models. Duplicate records must agree
exactly, and each candidate is matched separately, preventing a match
from spanning two candidate names. The archived files preserve parsed
pools, with incomplete raw-generation and serving-identifier records.
The collections differ in timing and serving provenance.

\begin{table*}[t]
\centering
\small
\begin{tabular}{@{}lrrr@{}}
\toprule
Model & Cars ($n=160$) & Laptops ($n=164$) & Headphones ($n=49$) \\
\midrule
GPT-5 & 80.0 [73.8, 86.2] & 89.0 [84.1, 93.9] & 69.4 [55.1, 81.6] \\
Claude Opus & 78.8 [71.9, 85.0] & 91.5 [87.2, 95.7] & 75.5 [63.3, 87.8] \\
Llama-70B & 75.6 [68.8, 81.9] & 95.7 [92.1, 98.2] & 57.1 [42.9, 69.4] \\
Qwen-80B & 72.5 [65.6, 79.4] & 92.7 [88.4, 96.3] & 59.2 [44.9, 71.4] \\
Nemotron-49B & 73.8 [66.9, 80.6] & 93.3 [89.0, 97.0] & 49.0 [34.7, 63.3] \\
Llama-8B & 70.0 [63.1, 76.9] & 92.1 [87.8, 96.3] & 53.1 [38.8, 65.4] \\
\bottomrule
\end{tabular}
\caption{\textbf{Recovered six-model reference coverage at $K=40$.}
Percentages with 95\% query-bootstrap intervals on shared request IDs
within each domain. All models use the same raw reference lists and
candidate-wise matcher. $K$ counts parsed entries, which can include
aliases, configurations, or malformed list items. No Mistral pools are
available for this comparison.}
\label{tab:coverage-six}
\end{table*}

Laptop coverage exceeds headphone coverage for all six models.
The highest coverage point estimates belong to GPT-5 for cars, Llama-70B for
laptops, and Claude for headphones.

\paragraph{Reference-filtering sensitivity.}
Inspection of the saved references reveals fragments such as
``Hyundai has'' and ``Sony is,'' alongside legitimate names and broad
families. As a symmetric sensitivity check, we retain only references
matching the existing purged catalog, applying one rule to every model.
At least one reference survives for 146/160 car, 160/164 laptop,
and 43/49 headphone requests. We keep all original queries in the
denominator, assigning zero coverage when no reference survives.
Table~\ref{tab:coverage-filtered} reports the resulting rates.
The reduction is largest for headphones, at 12.2--18.4 percentage
points across models. The filter can remove valid unlisted names,
and family-level matches remain. Reference endorsement and suitability
require independent validation.

\begin{table}[t]
\centering
\small
\begin{tabular}{@{}lrrr@{}}
\toprule
Model & Cars & Laptops & Headphones \\
\midrule
GPT-5 & 76.9 & 88.4 & 57.1 \\
Claude Opus & 77.5 & 91.5 & 61.2 \\
Llama-70B & 71.2 & 95.7 & 40.8 \\
Qwen-80B & 68.8 & 92.1 & 40.8 \\
Nemotron-49B & 68.1 & 92.7 & 34.7 \\
Llama-8B & 67.5 & 92.1 & 38.8 \\
\bottomrule
\end{tabular}
\caption{\textbf{Catalog-filtered reference coverage at $K=40$.}
Percentages on the unchanged 160/164/49 denominators from
Table~\ref{tab:coverage-six}. Only the reference lists change;
generated pools and matching rules are fixed.}
\label{tab:coverage-filtered}
\end{table}

\paragraph{Generated-pool choice across models.}
An archived GPT car chooser run covers all 171 car requests;
we verify its displayed pools against the saved enumeration records.
GPT-5.4 enumeration and choice on the same
171 car and 49 headphone requests uses the same prompts, seeds,
token limits, and first-mention pool order, with 1,320
calls. Both the historical number parser and a strict first-line
parser agree for every GPT-5.4 selection. Table~\ref{tab:generated-choice-extension}
distinguishes the archive from the October 2026 collection. GPT-5.4 coverage
rates are 80.7\% and 65.3\%, but reference selection conditional on
coverage is 30.9\% and 24.0\%. Conditional rates use each model's own
generated pools.

GPT-5.4 laptop enumeration and choice uses 1,056 calls on
the same 176 requests. Coverage is 159/176 (90.3\%), with
reference selection in 119/477 covered choice slots (24.9\%).
All 528 choices agree under the historical and strict parsers.
The same catalog filter gives 157/176 coverage (89.2\%) and
119/471 conditional selections (25.3\%). Broad family references
and unconstrained product dates remain; these rates describe
reference agreement on model-specific pools.
Opus enumeration and choice uses 294 calls on the same
49 headphone requests, covering 37/49 (75.5\%) and selecting a
reference in 39/111 covered choice slots (35.1\%). All 147 choices
agree under both parsers; source/identity sensitivities follow below.

\begin{table*}[t]
\centering
\small
\begin{tabular}{@{}llrrrr@{}}
\toprule
Domain & Model / collection & Queries & Coverage@40 & Covered & Selection when covered \\
\midrule
Cars & Llama-70B (archive) & 171 & 76.6 [70.2, 83.0] & 131 & 26.5 [18.8, 34.4] \\
Cars & GPT-5 (archive) & 171 & 80.7 [74.9, 86.0] & 138 & 33.3 [25.6, 41.6] \\
Cars & GPT-5.4 (Oct. 2026) & 171 & 80.7 [74.9, 86.0] & 138 & 30.9 [23.4, 37.9] \\
Headphones & Llama-70B (archive) & 49 & 57.1 [42.9, 69.4] & 28 & 39.3 [22.6, 57.1] \\
Headphones & GPT-5.4 (Oct. 2026) & 49 & 65.3 [53.1, 77.6] & 32 & 24.0 [9.4, 38.5] \\
Laptops & Llama-70B (archive) & 176 & 95.5 [92.0, 98.3] & 168 & 47.0 [40.1, 54.6] \\
Laptops & GPT-5.4 (Oct. 2026) & 176 & 90.3 [85.8, 94.3] & 159 & 24.9 [18.9, 31.7] \\
Headphones & Opus 4.8 (Oct. 2026) & 49 & 75.5 [63.3, 87.8] & 37 & 35.1 [18.9, 51.4] \\
\bottomrule
\end{tabular}
\caption{\textbf{Coverage and generated-pool selection across models.}
Rates are percentages with 95\% query-bootstrap intervals; Covered
counts requests whose pool contains a reference. Selection averages
three choices per covered request. The dated collection was gathered
October 10--12, 2026; archive denotes saved outputs from the
enumeration corpus. All collections share request IDs within a domain.}
\label{tab:generated-choice-extension}
\end{table*}

Applying the same catalog filter to the October 2026 GPT-5.4 collection changes coverage
to 78.9\% for cars and 53.1\% for headphones, with conditional
selection of 30.9\% (135 covered requests) and 25.6\% (26 requests).
Pools from either collection can contain products introduced after
the original posting date because the prompts leave availability unconstrained.

\paragraph{Headphone source and identity sensitivities.}
An assistant audit traced all 197 stored references on the same
49 headphone requests to 101 complete source comments. It labeled
52 exact endorsements, 18 conditional endorsements, 24 family-level
endorsements, 30 unendorsed mentions, 71 fragments/nonproducts, and
two unresolved identities. Exact endorsements remain for 28 queries;
the broad rule retaining exact, conditional, and family-level
endorsements covers 40. Labels were frozen before per-model scoring.
Missing or filtered references remain unassessable source cases,
not objective model failures.

\begin{table*}[t]
\centering
\small
\setlength{\tabcolsep}{4pt}
\begin{tabular}{@{}lrrrrr@{}}
\toprule
Model & Raw /49 & Exact /49 & Broad /49 & Broad /40 (\%) & Laptop gap (pp) \\
\midrule
GPT-5 & 34 & 12 & 24 & 60.0 [44.4, 74.4] & 29.0 [13.7, 45.5] \\
Claude Opus & 37 & 17 & 29 & 72.5 [57.9, 85.7] & 19.0 [5.1, 34.0] \\
Llama-70B & 28 & 8 & 20 & 50.0 [35.3, 65.8] & 45.7 [29.8, 60.6] \\
Qwen-80B & 29 & 7 & 20 & 50.0 [34.2, 65.0] & 42.7 [26.9, 58.9] \\
Nemotron-49B & 24 & 7 & 19 & 47.5 [32.4, 63.4] & 45.8 [29.5, 61.4] \\
Llama-8B & 26 & 8 & 19 & 47.5 [32.5, 63.4] & 44.6 [27.7, 59.8] \\
\bottomrule
\end{tabular}
\caption{\textbf{Headphone coverage under source-reference filtering.}
The first three numeric columns count covered queries among the
same 49. Broad /40 is coverage on source-eligible queries, with
95\% query-bootstrap intervals; Laptop gap subtracts it from raw
laptop coverage on 164 requests. Intervals use 4,000 resamples
(seed 20261010), independently sampling domains and pairing model
rows within each domain. Laptop references lack an equivalent
source audit.}
\label{tab:headphone-source-coverage}
\end{table*}

With the unchanged matcher, eligible-only broad coverage is
47.5--72.5\%, below laptop coverage for all six models
(Table~\ref{tab:headphone-source-coverage}). The individual intervals
for the domain gaps exclude zero, but omit annotation uncertainty
and are not multiplicity-adjusted. A separate blinded review of
153 historical positive identity matches found 62 supported,
22 unresolved, and 69 rejected matches. Restricting broad coverage
to supported identities yields 15, 20, 13, 12, 10, and 11 covered
queries out of 49 in the table's model order. This is a veto
sensitivity of existing positive matches; absent aliases and
unextracted source recommendations are not recovered.

\begin{table*}[t]
\centering
\small
\setlength{\tabcolsep}{4pt}
\begin{tabular}{@{}llrrr@{}}
\toprule
Run & Reference rule & Coverage & Selected / covered slots & Selection (\%; 95\% CI) \\
\midrule
Llama-70B & Raw & 28/49 & 33/84 & 39.3 [22.2, 57.0] \\
Llama-70B & Exact & 8/49 & 6/24 & 25.0 [0.0, 60.0] \\
Llama-70B & Broad & 20/49 & 20/60 & 33.3 [13.3, 54.3] \\
Llama-70B & Broad + identity & 13/49 & 9/39 & 23.1 [0.0, 50.0] \\
GPT-5.4 & Raw & 32/49 & 23/96 & 24.0 [9.9, 39.4] \\
GPT-5.4 & Exact & 14/49 & 6/42 & 14.3 [0.0, 35.7] \\
GPT-5.4 & Broad & 26/49 & 6/78 & 7.7 [0.0, 20.0] \\
GPT-5.4 & Broad + identity & 16/49 & 6/48 & 12.5 [0.0, 31.3] \\
Opus 4.8 & Raw & 37/49 & 39/111 & 35.1 [20.0, 51.3] \\
Opus 4.8 & Exact & 15/49 & 6/45 & 13.3 [0.0, 33.3] \\
Opus 4.8 & Broad & 28/49 & 21/84 & 25.0 [10.3, 42.3] \\
Opus 4.8 & Broad + identity & 17/49 & 3/51 & 5.9 [0.0, 20.0] \\
\bottomrule
\end{tabular}
\caption{\textbf{Headphone selection under the same source-reference rules.}
Each run retains 49 queries and three recorded choices per query.
Conditional selection divides matching choices by three times that
run's covered-query count. Intervals use 4,000 paired query-cluster
resamples (seed 20261010), recomputing the denominator each time.
Exact and broad references survive on 28 and 40 queries;
\emph{Broad + identity} additionally requires a supported positive match.}
\label{tab:headphone-source-selection}
\end{table*}

The same source rules also change conditional selection
(Table~\ref{tab:headphone-source-selection}). We use Llama-70B's
archived displayed pools and GPT-5.4's October 2026 pools from
Table~\ref{tab:generated-choice-extension}, not the GPT archive in the
six-model coverage table. Three choices remain per query; source
rules change both matching choices and covered denominators.
These ratios are sensitivities, not ordered bounds or model-quality
rankings. The raw-reference rows reproduce the original results.
Applying the identical source labels and identity rules to Opus
gives 28/49 broad-reference coverage and 21/84 conditional
selections (25.0\%). Requiring supported identity matches leaves
17/49 covered requests and 3/51 matching choices (5.9\%);
four name pairs were reviewed before aggregation.

The stored references are incomplete even after filtering. For one
request, all three GPT choices name Sony WH-1000XM2, which a source
comment endorses, but the stored complete reference names only
WH-1000XM3. The fixed no-addition filter therefore leaves genuine
source-endorsed choices uncounted. No independent human annotation,
complete alias recovery, or suitability judgment is supplied by this
audit. Llama's archive contains 147 valid picks but lacks historical
raw responses and transport records; GPT's 294 successful enumeration
and choice calls retain full logs. The Opus run also retains all 294 prompts and responses and underwent a fixed
ten-query raw audit and supplemental identity review. Collection date and
model-specific pools remain different across runs.

\paragraph{Hardware coverage and selection audit.}
We recover 234 archived Llama-70B Hardware Recommendations queries,
the complete predetermined subset selected by the original hash-bucket
rule from 1,266 reference-bearing questions. Every chooser pool matches
its saved enumeration prefix. The archive retains 701 of 702 planned
choices; the missing choice contributes zero. A generative assistant
reviewed all 296 reference entries against their full source answers,
independently of pool and choice outcomes. Of these, 235 identify an
endorsed specific product, 49 name families or ranges, and 12 are
unendorsed, truncated, compound, or unresolved. These annotations
assess source endorsement, not product suitability, and are not
independent human gold.

On the 194 queries with a specific-product reference, conservative
name-equivalence matching gives coverage of 21/194 (10.8\%) at
$K=10$ and 36/194 (18.6\%) at $K=40$. Among the 36 covered queries,
31/108 planned choices select a reference (28.7\%); excluding the
missing choice gives 31/107 (29.0\%).
Table~\ref{tab:hardware-recovery} separates exact lexical matching,
audited name equivalence, source-explicit identifiers, and family-level
credit. Across all stored references, the historical liberal matcher
instead gives 64/234 coverage (27.4\%), illustrating why reference
quality and identity rules matter.

\begin{table*}[t]
\centering
\small
\begin{tabular}{@{}lrr@{}}
\toprule
Reference and identity rule & Coverage at $K=40$ & Conditional selection \\
\midrule
Specific; strict lexical & 22/194: 11.3 [7.1, 16.2] & 19/66: 28.8 [12.0, 49.3] \\
Specific; audited equivalence & 36/194: 18.6 [13.1, 24.5] & 31/108: 28.7 [15.1, 44.4] \\
Specific; source-explicit identifiers & 37/194: 19.1 [13.6, 25.0] & 34/111: 30.6 [17.2, 45.9] \\
Specific or named family & 51/226: 22.6 [17.4, 28.1] & 45/153: 29.4 [18.2, 42.5] \\
\bottomrule
\end{tabular}
\caption{\textbf{Audited Llama-70B hardware coverage and selection.}
Cells give counts, percentages, and 95\% query-bootstrap intervals.
Selection uses three planned choices per covered query. Reference
eligibility and identity rules differ across rows; family credit
does not establish exact-product identity.}
\label{tab:hardware-recovery}
\end{table*}

This single-model subset provides descriptive reference agreement,
not a controlled domain comparison or an objective failure rate.
Some requests concern components, reuse, or software rather than a
new hardware purchase, and some already name products. Candidate
existence, historical availability, and complete configurations were
not validated. Raw generations and historical serving identifiers
are unavailable, preventing an independent choice-parser audit.
Intervals use 2,000 query-cluster resamples with seed 42, retaining
choices together and recomputing the covered denominator. They exclude
reference and matcher uncertainty.

\paragraph{GPT-5.4 collection audit.}
The October 2026 GPT-5.4 runs preserve every request payload, raw response, and
returned model identifier. We inspected the first ten frozen
cases in each coverage run, including all enumerations and choices,
and sampled archived pools across the six models and three domains.
The inspection was performed by a generative assistant.
We verified the prompts, shared request IDs, reference lists, and parser
agreement before calculating scores and catalog sensitivities. Intervals use 2,000 query resamples with seed 42, keeping all three
choices together within each request.

\paragraph{Enumeration and choice protocol.}
Enumeration requested three lists at temperature 0.9, nominal seeds 17--19,
and a 500-token output limit, requesting 20 names per list. The line parser
removes leading bullets or numbering and surrounding spaces, asterisks, and
backticks, then discards text after a colon or a whitespace-delimited dash.
It retains names of 3--59 characters unless they begin with here, sure, or note,
taking at most 25 names per response. The three lists are concatenated in sample
order and deduplicated case-insensitively, retaining first occurrences. Actual
list lengths and identities can differ from the requested 20 distinct names.
The chooser receives up to the first 40 names, with three choices at temperature
0.7, nominal seeds 42--44, and a 64-token output limit. It maps the first integer
in the response to the corresponding candidate if that integer is in range;
otherwise the choice is invalid. Hosted endpoints may not guarantee deterministic seeds.

The paired reanalysis intersects saved Llama-70B chooser records with
the free-form evaluation IDs, yielding 171/176/49 queries. Their
reference lists agree exactly across the two artifacts. The mean
displayed pool sizes are 36.4/36.2/34.8. A pool shorter than K is
evaluated in full, and a failed or invalid choice contributes zero
to the three-sample mean. We score each selected name directly against
the same raw reference set used for the free-form output. Numerical
choices have a different parsing interface from free text. The
extraction asymmetry documented below contributes to the measured score difference.

\begin{table*}[t]
\centering
\small
\begin{tabular}{@{}lrrrrrr@{}}
\toprule
Domain & $n$ & Coverage@10 & Coverage@40 & Free form & Enum. + choice & Change [95\% CI] \\
\midrule
Cars & 171 & 59.6 & 76.6 & 25.3 & 20.3 & $-5.1$ [$-10.9$, $+0.8$] \\
Laptops & 176 & 85.8 & 95.5 & 28.0 & 44.9 & $+16.9$ [$+7.6$, $+26.1$] \\
Headphones & 49 & 49.0 & 57.1 & 22.4 & 22.4 & $0.0$ [$-10.9$, $+12.2$] \\
\bottomrule
\end{tabular}
\caption{\textbf{Candidate coverage and observed final-choice scores.}
Llama-70B on matched queries; all values except $n$ are percentages or
percentage-point differences. Coverage@$K$ measures reference presence
among the first $K$ enumerated options. Both final-choice scores
average three samples; failed choices score zero. Intervals on the
paired change are 95\% query-bootstrap intervals. Additional
enumeration calls and different answer-extraction interfaces prevent
interpreting the change as an isolated improvement in recommendation.}
\label{tab:enum-protocol-comparison}
\end{table*}

\paragraph{Free-form extraction sensitivity.}
The saved laptop protocols differ by 16.9 percentage points
(Table~\ref{tab:enum-protocol-comparison}), but
that difference is not reliable evidence that enumeration improves
the recommendation. The free-text extractor recovers a product in
only 75.0\% of laptop samples; all saved numbered choices in this
matched comparison are valid. Moreover, 84 of the 89 net additional
matched samples occur on the 54 requests with at least one
unextractable free-form answer. An explicit ``Dell XPS 13'' can fail
catalog extraction while receiving credit when supplied directly as
a chooser's selected name. Thus a recognizable recommendation can
be penalized on only one side of the comparison.

Among the other 122 laptop requests, where all three free-form answers
are extractable, the observed gap is 1.4 points, with a 95\%
query-bootstrap interval of $[-9.6,12.3]$. This selected subset is a
sensitivity check, not a corrected estimate of the full-sample effect.
A shared, validated identity-extraction procedure is needed
before interpreting the full-sample difference as a gain.

\paragraph{Within-pool selection controls.}
Uniformly selecting a displayed slot gives mean reference agreement
of 7.8\% on cars, 11.6\% on laptops, and 3.2\% on headphones under
these exact pools and matching rules. The LLM chooser's point
estimates are higher in all three domains. These task-specific
baselines count every matching slot.

A stronger within-pool baseline selects the first enumerated name,
without an additional chooser call. It obtains 25.1\% on cars,
36.4\% on laptops, and 16.3\% on headphones. Chooser-minus-first
differences are respectively $-4.9$, $+8.5$, and $+6.1$ points, with
95\% paired query-bootstrap intervals $[-11.7,1.9]$,
$[-0.2,17.0]$, and $[-6.8,19.0]$. All three intervals include zero.
Both sides use direct product-name matching, so this comparison
avoids the free-text catalog gate. The first option itself comes
from a query-conditioned generation.

The sensitivity calculations use 10,000 paired query-bootstrap resamples with
seeds derived from stable hashes. They reuse the same saved responses without
new model calls.

The request body supplied to enumeration and choice is truncated at
1,500 characters, compared with 1,800 for free-form generation. A
cost- and format-controlled follow-up should use identical query
content, equal inference budgets, and the same answer interface.
The present comparison measures the full saved protocols.

\paragraph{Teacher-forced likelihood lift.}
Factual question-answering studies link recall to entity popularity
and pretraining support~\citep{mallen2023popqa,kandpal2023longtail}.
Our product-name scores probe a related distinction, but do not
establish whether a model knows a product's attributes.
The archived open-weight experiment scores a reference name
as a continuation. For answer tokens $a_1,\ldots,a_T$, define
$\ell(a\mid x)=T^{-1}\sum_{t=1}^T\log p(a_t\mid x,a_{<t})$.
Lift is $\ell(a\mid x_q)-\ell(a\mid x_0)$, where $x_q$ includes
the request and $x_0$ does not. Both conditions score the identical
token sequence under the model's chat template. The car prompts are:
\begin{quote}\small
Without request: ``Recommend one specific car. Reply with
the product name only.''

With request: [request text], followed by ``Recommend one
specific car for this request. Reply with the product name only.''
\end{quote}
The product category changes with the domain. The request is
the title and body truncated jointly to 900 characters. The reference
is the first stored name recognized by the purged catalog or Stack
Exchange lexicon. Evaluation uses bfloat16 and teacher-forces answer
tokens without scoring an end-of-sequence token. The model identifiers are
\path{meta-llama/Llama-3.1-8B-Instruct},
\path{unsloth/Llama-3.3-70B-Instruct}, and
\path{Qwen/Qwen3-Next-80B-A3B-Instruct}.

\begin{table*}[t]
\centering
\small
\begin{tabular}{@{}llrrrr@{}}
\toprule
Domain & Model & Queries & Median lift [95\% CI] & Lift $>0$ (\%) & Lift $>0.5$ (\%) \\
\midrule
Cars & Llama-8B & 1,292 & 1.21 [1.09, 1.34] & 76.9 & 66.0 \\
Cars & Llama-70B & 1,292 & 0.44 [0.21, 0.77] & 55.3 & 49.5 \\
Cars & Qwen-80B & 1,292 & 1.61 [1.42, 1.77] & 73.2 & 67.0 \\
\midrule
Laptops & Llama-8B & 902 & 0.85 [0.75, 0.94] & 74.8 & 61.5 \\
Laptops & Llama-70B & 902 & 1.42 [1.19, 1.70] & 68.5 & 60.1 \\
Laptops & Qwen-80B & 902 & 1.84 [1.66, 2.00] & 83.6 & 73.2 \\
\midrule
Headphones & Llama-8B & 198 & 1.02 [0.83, 1.22] & 79.8 & 65.7 \\
Headphones & Llama-70B & 198 & 0.83 [0.26, 1.08] & 59.1 & 53.0 \\
Headphones & Qwen-80B & 198 & 1.16 [1.02, 1.37] & 81.3 & 73.7 \\
\midrule
SE software & Llama-8B & 2,445 & 3.24 [3.14, 3.36] & 93.7 & 90.0 \\
SE software & Llama-70B & 2,445 & 3.82 [3.64, 3.98] & 84.6 & 81.4 \\
SE software & Qwen-80B & 2,445 & 6.25 [6.04, 6.44] & 97.1 & 95.9 \\
\midrule
SE hardware & Llama-8B & 1,266 & 1.49 [1.41, 1.56] & 92.4 & 82.5 \\
SE hardware & Llama-70B & 1,266 & 1.96 [1.74, 2.13] & 80.6 & 74.4 \\
SE hardware & Qwen-80B & 1,266 & 2.19 [2.08, 2.30] & 93.4 & 87.6 \\
\bottomrule
\end{tabular}
\caption{\textbf{Effect of the request-specific prompt on reference-name likelihood.}
Lift is the paired change in mean log-probability, in nats per answer
token. Medians retain the archived upper-median convention, with 95\%
intervals from 2,000 query-bootstrap resamples. Fractions are
descriptive counts over the same queries; the 0.5 threshold is the
original probe's summary threshold. Positive lift indicates increased
name likelihood.}
\label{tab:likelihood-lift}
\end{table*}

Table~\ref{tab:likelihood-lift} reports all three models
across five domains. The existing cluster runs were recovered without
new inference. We checked that every saved answer matches the corresponding input and verified
the paired score differences. Intervals quantify variation across requests,
conditional on the archived model scores rounded to three decimals;
they do not include checkpoint or numerical-precision variability.

This probe measures prompt-conditioned name likelihood. Inferring
product knowledge would require additional evidence about the model's
knowledge of its attributes. Both the request and the phrase ``for this request'' change,
so lift is not an isolated effect of the request text. References may
already appear in the request, and a likelihood decrease can reflect
an unsuitable reference. The probe does not rank all competing products
or reproduce multi-item enumeration.

\paragraph{Linking likelihood lift to pool omissions.}
We joined the saved Llama-70B car scores to enumeration pools by
exact reconstructed request text and query ID. The primary cohort
is the 156-request overlap with the fixed 171-query enumeration
sample; 15 requests lack a score under the historical lift
eligibility rule. The target is the first catalog-valid stored
reference, which is the first raw reference for 102/156 requests.
Target omission therefore differs from omission of every stored
reference. Candidate-wise matching at $K=40$ reproduces all 156
existing any-reference coverage labels.

\begin{table*}[t]
\centering
\small
\begin{tabular}{@{}lrrr@{}}
\toprule
Pool outcome & Queries & Mean lift [95\% CI] & Positive lift (\%) \\
\midrule
Target present & 87 & $1.75\;[0.87, 2.66]$ & 58 / 87 (66.7) \\
Target absent & 69 & $-0.52\;[-1.42, 0.32]$ & 28 / 69 (40.6) \\
All references absent (subset) & 25 & $-0.76\;[-1.95, 0.49]$ & 9 / 25 (36.0) \\
\bottomrule
\end{tabular}
\caption{\textbf{Reference-name likelihood lift and explicit pool inclusion.}
Llama-70B artifacts on 156 matched car requests, with coverage at
$K=40$. The last row is a subset of target-absent requests. Lift is
measured in nats per target token; mean intervals use 2,000
query-bootstrap resamples (seed 20261010). The final column gives
counts and percentages with positive lift. Scores and pools come
from different prompts and serving implementations.}
\label{tab:lift-coverage-join}
\end{table*}

Included targets have 2.28 more nats/token mean lift than omitted
targets (95\% CI 1.06--3.56; Table~\ref{tab:lift-coverage-join}).
Yet positive lift occurs for 28/69 omitted targets, including 9/25
requests where every stored reference is absent. The association
also appears among 74 requests with complete inputs within the
900-character limit and no full target-name mention in either
prompt (difference 2.52, CI 0.87--4.19), and in the separate
577-request archive overlap (2.09, CI 1.44--2.72).

The protocol fixed these cohorts and sensitivity checks before
computing the association. Assistant review covered 12 fixed raw
cases and all 67 disagreements with exact surface matching.
Some references are incidental or critical mentions. No
outcome-dependent reference cleaning was applied. The local scorer
is \path{unsloth/Llama-3.3-70B-Instruct}; the historical pool-serving
alias lacks a common checkpoint fingerprint. Scoring truncates
requests at 900 characters, versus title plus 1,500 body characters
for enumeration. These differences make the result a descriptive
association between contextual name likelihood and pool inclusion;
it does not establish product knowledge, suitability, or a causal
explanation of omissions.

\section{Query-conditioned choice details}
\label{app:lift}

Table~\ref{tab:lift} summarizes supplied-choice results, including
the separate ESCI test described in Appendix~\ref{app:esci}.

\begin{table*}[t]
\centering
\small
\begin{tabular}{@{}lrrrrrr@{}}
\toprule
& \multicolumn{5}{c}{Community reference} & Relevance label \\
\cmidrule(lr){2-6}\cmidrule(l){7-7}
Model & Cars & Laptops & Headphones & SE software & SE hardware & ESCI \\
\midrule
GPT-5 & 59.4 & 54.0 & 54.0 & 96.7 & 94.6 & 70.2 \\
Claude Opus & 62.5 & 58.5 & 51.5 & 96.3 & 94.4 & 73.6 \\
Llama-70B & 57.0 & 60.5 & 51.0 & 94.4 & 91.3 & 67.4 \\
Llama-8B & 56.9 & 51.1 & 46.0 & 93.7 & 92.3 & 58.6 \\
\midrule
Metadata control & 86.2 & 64.3 & 88.4 & 97.3 & 97.9 & --- \\
Queries & 1,292 & 902 & 198 & 2,445 & 1,266 & 500 \\
\bottomrule
\end{tabular}
\caption{\textbf{Selection with the reference supplied.} Percentage choosing
the community reference with the user's request (five options), or
the Exact-labeled ESCI product (four options). Models share queries
within each column. The metadata control uses reconstructed popularity
ranks unavailable in the model prompts. The two tasks use different
reference criteria and candidate pools.}
\label{tab:lift}
\end{table*}

The payload selects the first reference string recognizable by the
purged catalog vocabulary (Reddit) or product lexicon (Stack Exchange).
Query text is the
title and body truncated jointly to 900 characters. Distractors are
the first four nonmatching names among the first 60 entries of a
popularity-ordered list. Options are shuffled once per query and kept
fixed across the two prompt conditions within that run.

Both prompt conditions use temperature zero and a 200-token response budget.
Parsing first seeks an explicit \texttt{CHOICE: <number>}, case-insensitively
and allowing an optional \texttt{\#} before the number. If that number is absent
or out of range, the first in-range integer anywhere in the response is used.
Invalid responses count as misses. Stored records contain
query text, the reference, its position, and two hit flags, but not
the complete option list or raw responses. The shuffle uses Python's
process-dependent string hash, so its nominal seed does not guarantee
identical positions across different model runs. We can reproduce the
reported hit rates and paired intervals; individual choice and parsing
decisions remain partly untraceable.

Table~\ref{tab:lift-both-conditions} reports both prompt
conditions across all five domains; Table~\ref{tab:lift-ci} gives
paired intervals for their differences. Reference selection increases
in all 20 model--domain cells.

\begin{table*}[t]
\centering
\small
\begin{tabular}{@{}lrrrrr@{}}
\toprule
Model & Cars & Laptops & Headphones & SE software & SE hardware \\
\midrule
GPT-5 & 5.0$\to$59.4 & 16.4$\to$54.0 & 37.9$\to$54.0 & 1.8$\to$96.7 & 2.4$\to$94.6 \\
Claude Opus & 7.4$\to$62.5 & 14.9$\to$58.5 & 33.3$\to$51.5 & 1.3$\to$96.3 & 6.2$\to$94.4 \\
Llama-70B & 7.8$\to$57.0 & 27.7$\to$60.5 & 28.8$\to$51.0 & 3.5$\to$94.4 & 16.1$\to$91.3 \\
Llama-8B & 9.0$\to$56.9 & 20.0$\to$51.1 & 18.2$\to$46.0 & 2.5$\to$93.7 & 10.0$\to$92.3 \\
\midrule
Queries & 1,292 & 902 & 198 & 2,445 & 1,266 \\
\bottomrule
\end{tabular}
\caption{\textbf{Query-conditioned choice in a supplied five-option pool.}
Percentage choosing the reference without $\to$ with the request.
The instruction changes from ``best in general'' to ``best for this
user.'' Models share query identities within each domain; options
are fixed across prompt conditions within each run.}
\label{tab:lift-both-conditions}
\end{table*}

\paragraph{A candidate-construction control.}
We reconstructed the unordered option sets using the saved references,
the current popularity-ordered list, and the four-distractor rule specified above. Reference identities and query sets agree across
all four models and with the payload. The control selects the option
with the largest index in the popularity list; it uses no request
text and no fitted parameters. Because the four distractors come from
the head of that list, this rule often identifies the inserted
reference. There are only five or six distinct distractor tuples per
domain. The control is invariant to option order, but the archived
records do not establish that every historical option list exactly
matches this reconstruction. This diagnostic is analogous to
hypothesis-only checks for annotation artifacts in natural-language
inference~\citep{gururangan2018annotation}: it tests whether
construction cues support prediction without the intended input.

\begin{table*}[t]
\centering
\small
\begin{tabular}{@{}lrrr@{}}
\toprule
Domain & Rank control (\%) & Best model (\%) & Difference [95\% CI] \\
\midrule
Cars & 86.2 & 62.5 & +23.7 [+20.7, +26.7] \\
Laptops & 64.3 & 60.5 & +3.8 [+0.2, +7.3] \\
Headphones & 88.4 & 54.0 & +34.3 [+26.3, +42.4] \\
SE software & 97.3 & 96.7 & +0.6 [-0.3, +1.6] \\
SE hardware & 97.9 & 94.6 & +3.3 [+2.0, +4.7] \\
\bottomrule
\end{tabular}
\caption{\textbf{A query-free shortcut in reconstructed choice sets.}
Reference-selection rates for the least-popular-option control and
the highest query-conditioned model point estimate in each domain.
Differences have 95\% paired query-bootstrap intervals over the saved
query sets (10,000 resamples). The control uses construction metadata
unavailable in the model prompts. Intervals are exploratory and
not corrected for selecting the best model or multiple comparisons.}
\label{tab:shortcut}
\end{table*}

Table~\ref{tab:shortcut} rules out treating the high conditioned hit
rates as sufficient evidence of distinguishing subtle preferences.
It does not establish that the models used the same shortcut. A
separate token-overlap control, choosing the option sharing most
normalized tokens with the request and averaging uniformly over
ties, scores only 31.3\%, 30.3\%, 34.5\%, 39.5\%, and 54.6\%
in the table's domain order. Thus this simple lexical rule does not
explain the models' high rates either.

The historical prompt comparison changes both the instruction and
the request, with the reference supplied in both conditions.
The balanced-pair test holds the instruction fixed and varies
request content on frequency-balanced candidate pairs.

\begin{table*}[t]
\centering
\small
\setlength{\tabcolsep}{4pt}
\begin{tabular}{@{}lrrrr@{}}
\toprule
Domain & GPT-5 & Claude Opus & Llama-70B & Llama-8B \\
\midrule
Cars & 54.4 [51.5, 57.5] & 55.2 [52.2, 58.2] & 49.2 [46.1, 52.2] & 47.9 [44.9, 51.0] \\
Laptops & 37.6 [33.9, 41.5] & 43.7 [39.6, 47.6] & 32.8 [28.5, 36.9] & 31.2 [27.2, 35.3] \\
Headphones & 16.2 [7.1, 25.8] & 18.2 [10.1, 26.8] & 22.2 [13.1, 31.3] & 27.8 [19.2, 35.9] \\
SE software & 94.9 [93.9, 95.8] & 95.0 [94.1, 95.8] & 90.9 [89.7, 92.0] & 91.1 [89.9, 92.4] \\
SE hardware & 92.3 [90.8, 93.7] & 88.2 [86.3, 89.9] & 75.2 [72.7, 77.6] & 82.3 [80.2, 84.5] \\
\bottomrule
\end{tabular}
\caption{\textbf{Paired query-lift estimates.} Percentage-point change
from query-blind to conditioned selection, with 95\% query-bootstrap
intervals. All estimates use paired saved hit flags. They do not
measure uncertainty in labels, distractor construction, or parsing.}
\label{tab:lift-ci}
\end{table*}

\paragraph{Balanced-pair request control.}
We test 94 car requests with GPT-5.4, Llama-70B,
Claude Opus 4.8, and Qwen-80B,
using the same instruction with the real request, no request,
or a deterministically shuffled request. Each query has one stored
reference and one alternative from its archived Llama candidate
pool, presented in both orders: 564 calls per model and 2,256 in total.
Pairs share
a coarse catalog category and have corpus-reference frequencies
within a factor of two. The reference is more frequent in 47 pairs
and less frequent in 47, so a frequency-only rule selects it 50\%
of the time. Six source-only category exclusions from an initial
100-query sample were fixed before outcomes, without replacements.
Generic model-family options and source mentions do not establish
verified suitability.

\begin{table*}[t]
\centering
\small
\setlength{\tabcolsep}{4pt}
\begin{tabular}{@{}llrrrrrr@{}}
\toprule
Model & Request & Reference & Alternative & Equal & Neither & Insufficient & Invalid \\
\midrule
GPT-5.4 & Real & 86 & 65 & 14 & 21 & 2 & 0 \\
GPT-5.4 & Absent & 0 & 0 & 0 & 0 & 188 & 0 \\
GPT-5.4 & Shuffled & 29 & 27 & 10 & 121 & 1 & 0 \\
Llama-70B & Real & 55 & 45 & 17 & 18 & 52 & 1 \\
Llama-70B & Absent & 0 & 0 & 0 & 0 & 188 & 0 \\
Llama-70B & Shuffled & 11 & 13 & 5 & 107 & 48 & 4 \\
Opus 4.8 & Real & 84 & 49 & 27 & 17 & 11 & 0 \\
Opus 4.8 & Absent & 0 & 0 & 0 & 0 & 188 & 0 \\
Opus 4.8 & Shuffled & 25 & 26 & 10 & 120 & 7 & 0 \\
Qwen-80B & Real & 69 & 56 & 16 & 30 & 15 & 2 \\
Qwen-80B & Absent & 0 & 0 & 9 & 0 & 179 & 0 \\
Qwen-80B & Shuffled & 17 & 23 & 4 & 135 & 6 & 3 \\
\bottomrule
\end{tabular}
\caption{\textbf{Balanced-pair selection outcomes by request condition.}
Counts out of 188 presentations per row (94 requests, both option
orders), at temperature zero and seed 42. Equal, neither, and
insufficient-information responses are distinct from choosing an
option; invalid outputs remain in the denominator.}
\label{tab:controlled-selection}
\end{table*}

Real-request reference agreement is 45.7\% for GPT and 29.3\%
for Llama, versus 15.4\% and 5.9\% under shuffled requests
(Table~\ref{tab:controlled-selection}). Paired real-minus-shuffled
differences are $+30.3$ points (95\% CI $[19.1,41.5]$) and $+23.4$
points ($[13.3,34.0]$), using 5,000 whole-query bootstrap resamples
(seed 20261013). Opus gives 44.7\% agreement with real requests
and 13.3\% with shuffled requests, a paired increase of $31.4$
points ($[20.7,42.0]$). Qwen gives 36.7\% versus 9.0\%, an increase
of $27.7$ points ($[17.0,38.3]$). Without a request, GPT, Llama,
and Opus always report insufficient information; Qwen does so
179/188 times and reports equal suitability nine times.
Shuffling often changes the relevant product
category and elicits rejection of both options.

The models also differ in willingness to choose: real-request
decisive rates are 151/188 for GPT and 100/188 for Llama.
Conditional on choosing either option, reference agreement is
86/151 (57.0\%; 95\% CI $[46.2,67.4]$) and 55/100
(55.0\%; $[41.8,68.0]$), respectively. Thus, the unconditional
difference cannot establish a selection-accuracy advantage.
Opus chooses an option in 133/188 presentations and selects the
reference in 84/133 decisive choices (63.2\%; $[51.7,73.7]$).
Qwen selects the reference in 69/125 decisive choices
(55.2\%; $[43.6,66.9]$).
These outcomes measure request-sensitive choices, rejection, and abstention;
it retains coarse categories, imperfect source references, and
conditioning on an archived Llama pool.

All payloads, option orders, serving identifiers, and outcome mappings
were checked. Assistant review covered all 144 responses on a fixed
12-query audit and all five malformed outputs, followed by all
72 responses each for Opus and Qwen on the same audit queries and all
five malformed Qwen outputs. A symmetric parser
accepts one whole outer JSON fence; malformed content is not repaired.
The initial calls use
\path{openai/openai/gpt-5.4} and
\path{nvidia/meta/llama-3.3-70b-instruct}; an earlier route attempt failed. The Opus collection
uses \path{azure/anthropic/claude-opus-4-8} with unchanged pairs,
donor requests, and scoring. Two Opus responses require outer-fence
removal; both report insufficient information without a request.
The Qwen collection
uses \path{nvidia/qwen/qwen3-next-80b-a3b-instruct}. Its five
malformed outputs contain unescaped quotes and remain invalid.

\paragraph{Separating product switches from non-selection.}
A retrospective reanalysis pairs each real-request outcome with its
shuffled-request outcome for the same query and option order. The
change in the reference-selection indicator partitions exactly into
three net contributions: switches between the alternative and the
reference; transitions between a valid non-selection
(equal, neither, or insufficient information) and the reference;
and transitions between an invalid output and the reference.
Each transition into the reference contributes $+1$, and its reverse
contributes $-1$, using all 188 presentations per model as the
denominator. This accounting retains non-selections and parser
failures, without conditioning on a changed set of decisive answers.

\begin{table*}[t]
\centering
\small
\begin{tabular}{@{}lrrrr@{}}
\toprule
Model & Total $\Delta$ & Product switch & Non-selection & Invalid \\
\midrule
GPT-5.4 & 30.3 [19.1, 41.5] & 1.6 [-3.2, 6.4] & 28.7 [18.6, 39.4] & 0.0 [0.0, 0.0] \\
Llama-70B & 23.4 [13.3, 34.0] & 0.5 [-1.6, 2.7] & 22.9 [13.3, 33.0] & 0.0 [0.0, 0.0] \\
Opus 4.8 & 31.4 [20.7, 42.0] & 4.3 [0.0, 9.0] & 27.1 [16.5, 37.2] & 0.0 [0.0, 0.0] \\
Qwen-80B & 27.7 [17.0, 38.3] & 0.0 [-4.8, 4.8] & 26.6 [18.1, 36.2] & 1.1 [0.0, 2.7] \\
\bottomrule
\end{tabular}
\caption{\textbf{Decomposition of real-minus-shuffled reference agreement.}
Net contributions in percentage points, with descriptive 95\% intervals
from 5,000 paired whole-query resamples (seed 20261013); each resample
retains both option orders and all models. The three contributions sum
to the total before rounding. Intervals are not adjusted for multiple
comparisons. Non-selection includes equal, neither, and insufficient;
invalid outputs remain in the denominator. These are reference-agreement
changes, not correctness scores or causal mediation estimates.}
\label{tab:selection-transitions}
\end{table*}

The net non-selection contribution is 22.9--28.7 percentage points,
whereas product-to-product switches contribute 0.0--4.3 points
(Table~\ref{tab:selection-transitions}). All product-switch intervals
include or touch zero. Net contributions can conceal changes in
both directions: Qwen switches from alternative to reference in
seven presentations and from reference to alternative in seven.
The overall agreement increase thus mainly reflects transitions
between non-selection and selecting the reference. It does not
demonstrate improved ranking among suitable products, especially
when shuffled requests change the relevant product category.

The analysis uses
all 2,256 saved responses. Validation reproduces the condition counts
and total-effect intervals. An assistant checks all
48 real/shuffled responses on three hash-selected requests and all
ten malformed responses. Declared choices determine the outcome;
incorrect product assertions in explanations are not treated as
evidence of suitability. The transition analysis neither adjudicates
those assertions nor provides independent human validation.

\section{ESCI selection protocol}
\label{app:esci}

We filter to US records with
\texttt{small\_version==1}, pooling the official train and test splits.
Queries must begin with a lowercase ASCII letter
or digit, contain two to six whitespace-separated tokens, and contain
neither \texttt{\#} nor \texttt{!}. Each retained query must have at
least one Exact product and at least three Irrelevant products with
available titles. These surface restrictions define the sampling frame.

We sample query IDs using Python's \texttt{random.Random(0)} over
sorted eligible IDs, then uses a query-ID seed to choose one Exact
and three Irrelevant products and shuffle their titles. Each title is
truncated to 120 characters. Models are prompted at temperature zero
with a 120-token output limit. The parser first seeks an explicit
\texttt{CHOICE: <number>}, then falls back to an in-range number.
An invalid selection is scored as a miss. The artifacts contain the
same 500 query IDs across models and respectively 368, 351, 337, and
293 hits for Claude, GPT-5, Llama-70B, and Llama-8B.
Qwen-80B uses the identical queries, candidate titles, ordering, and prompt
(collected October 10, 2026): 339/500 Exact-label matches
(67.8\%). All 500 outputs are valid under both the historical and
strict first-line parsers. This run retains raw responses and returned
model identifiers; the other four model records retain only hit flags.

\begin{table}[t]
\centering
\small
\begin{tabular}{@{}lrr@{}}
\toprule
Model & Hit rate (\%) & 95\% CI \\
\midrule
Claude Opus & \textbf{73.6} & [69.6, 77.3] \\
GPT-5 & 70.2 & [66.0, 74.0] \\
Llama-70B & 67.4 & [63.2, 71.4] \\
Qwen-80B & 67.8 & [63.6, 71.7] \\
Llama-8B & 58.6 & [54.2, 62.8] \\
\bottomrule
\end{tabular}
\caption{\textbf{Externally labeled supplied choice.} Exact-reference
hit rate on the same 500 ESCI queries, with 95\% Wilson intervals.
Uniform choice is 25\%. Higher means greater agreement with the
designated label.}
\label{tab:esci}
\end{table}

Table~\ref{tab:esci} reports intervals for each hit rate.
The point estimates span 15 percentage points across models, and
Llama-70B exceeds Llama-8B by 8.8 points. Claude and GPT-5 miss the Exact
product on 132/500 and 149/500 queries, respectively. The GPT, Claude, and two Llama
records retain hit flags rather than raw responses, preventing a retrospective
separation of reasoning errors, invalid outputs, title truncation,
and label noise. The Qwen run permits inspection of raw
choices. In its first ten frozen cases, the query ``helen butler''
has a license-plate lamp as its Exact-labeled option, illustrating
that not every miss can safely be called a selection error.

The larger Llama wins on 80 discordant queries and loses on 36 against
the smaller model; a two-sided exact paired test gives
$p\approx5.4\times10^{-5}$. The GPT-5--Llama-70B contrast has 47 wins
and 33 losses ($p\approx0.146$). These exploratory comparisons
provide evidence for the larger Llama's advantage, while the
GPT-5--Llama-70B difference remains inconclusive.

These results describe the specified sampling frame; they do not
establish that the surface filters causally reduce label noise.
Auditing the displayed products and preserving raw selections are
necessary to identify the remaining error sources.

\clearpage
\section{Artifact attribution and reproducibility details}
\label{app:artifacts}

\subsection{Artifact origins}
The Reddit requests and community answers originate from contributors to
r/whatcarshouldIbuy, r/SuggestALaptop, and r/HeadphoneAdvice. The Stack Exchange
material originates from contributors to Software Recommendations and Hardware
Recommendations, distributed through Stack Exchange's data dumps.\footnote{\url{https://archive.org/details/stackexchange}}
Section~\ref{sec:setup} and Appendix~\ref{app:coverage} document the cohorts
and filtering. The externally labeled shopping subset derives from Amazon's
Shopping Queries Dataset \citep{reddy2022esci}; Appendix~\ref{app:esci}
specifies its construction and evaluation scope.

The evaluated open-model families are credited to Meta \citep{meta2024llama31,meta2024llama33},
Qwen \citep{qwen2025next}, NVIDIA \citep{nvidia2025nemotron}, and Mistral
AI \citep{mistral2025ministral}. The hosted frontier models are from
OpenAI \citep{openai2026gpt54} and Anthropic \citep{anthropic2026opus48}.
These references identify the model creators; the requested serving identifiers
and their historical provenance limits are documented in Appendix~\ref{app:repro}.
In particular, the Mistral gateway alias differs from the official
\path{Ministral-3-14B-Instruct-2512} model-card name. The archived Llama-70B
likelihood probe used the Unsloth distribution
\path{unsloth/Llama-3.3-70B-Instruct}, which credits Meta's original model.\footnote{\url{https://huggingface.co/unsloth/Llama-3.3-70B-Instruct}}
This checkpoint should be distinguished from the hosted Llama route.

\subsection{Software and evaluation settings}
Preprocessing and evaluation use Python, NumPy \citep{harris2020numpy}, and
pandas \citep{mckinney2010pandas}, with Requests for HTTP clients and PyArrow
for Parquet data.\footnote{\url{https://www.python.org/}; \url{https://github.com/psf/requests}; \url{https://arrow.apache.org/docs/python/}}
The recorded full-request, open-model car, and temperature-zero client
records report Python 3.8.3, NumPy 1.24.4, pandas 1.0.5, and Requests 2.24.0.
The laptop coverage/selection client record reports pandas 1.5.3 and
Requests 2.32.4 with the same Python and NumPy versions. Both profiles record
PyArrow 16.1.0. The final headphone connection audit separately records
Python 3.8.3 and NumPy 1.24.4, with no GPU use. These are per-run client/analysis
settings; they do not describe the providers' model-serving environments.

Product normalization, alias matching, reference extraction, and choice
parsing follow the task-specific rules described in
Appendices~\ref{app:reanalysis-protocol}--\ref{app:agreement}, including
normalization scope, malformed-output handling, denominators, and uncertainty
estimation. Bootstrap resampling uses NumPy; the experiment-specific appendices
give the resampling unit, replicate count, and seed or deterministic seed
derivation. Prompts, generation limits, temperatures, and ordering controls
are reported in Appendix~\ref{app:repro} and the corresponding experiment
appendices. These are inference experiments, with no model training on the
evaluation queries.

The archived teacher-forced likelihood implementation uses PyTorch and Hugging
Face Transformers \citep{wolf2020transformers}.\footnote{\url{https://github.com/pytorch/pytorch}}
The model is loaded in \texttt{bfloat16} with automatic device placement and
evaluated without gradients, one sequence at a time. Prefixes use the
tokenizer's chat template with a generation prompt. Prefix and answer tokens
are encoded separately with \texttt{add\_special\_tokens=False}; the mean
answer-token log probability is computed using float32 log-softmax, and
archived scores are rounded to three decimals.

\subsection{Compute resources}
The open-model names indicate nominal scales of 8B and 70B parameters for
Llama, 80B total and 3B active for Qwen3-Next, 49B for Nemotron, and 14B for
Ministral. Exact parameter counts for the hosted GPT and Claude models are
unavailable in the retained service records. The recorded local client/analysis
host is Linux x86-64 with 20 logical CPUs and approximately 31 GiB of RAM.
Hosted generation runs use remote inference services; this host specification
does not describe their accelerators. The archived likelihood probes ran on
a Slurm-managed cluster.

The Ethical considerations section reports response counts and available
token usage for the listed hosted-inference runs, including the 826-response
temperature-zero check. These counts measure the evaluated workload; they are
not estimates of provider-side GPU-hours, energy consumption, or total project
compute.

\subsection{Human involvement, data release, and AI assistance}
No human annotators were recruited and no participant study was conducted.
The experiments reuse public posts and saved model outputs. Source and response
reviews described as assistant audits are AI-assisted checks, not independent
human annotations. Public availability of the underlying posts does not
establish individual consent for every downstream use; we make no claim of
individual consent or institutional ethics-board approval or exemption.

No dataset, raw user-text collection, or model weights are being released with
this submission. This release status does not certify that the working data
are anonymized or cleared for redistribution. Any future data release requires
the privacy, licensing, attribution, and removal-process review, including
screening for identifying and potentially objectionable content. The present
study does not establish demographic representativeness beyond the platform,
language, domain, and sampling scope documented in Section~\ref{sec:setup},
Appendix~\ref{app:coverage}, and Limitations.

Generative AI assistants provided substantive support for experimental code,
analysis and source checks, literature lookup, and manuscript drafting and
revision. This assistance is distinct from evaluating LLMs as research objects.
The authors are responsible for verifying the final evidence, citations,
claims, and disclosures.

\end{document}